\pdfoutput=1
\documentclass[11pt,a4paper]{article}
\usepackage[hyperref]{emnlp-ijcnlp-2019}
\usepackage{times}
\usepackage{latexsym}

\usepackage{url}

\usepackage{microtype} %
\usepackage{amsmath}
\usepackage{booktabs}
\usepackage{wasysym}
\usepackage{xcolor}
\usepackage{tikz}
\usetikzlibrary{tikzmark, decorations.markings, calc, bending, patterns, positioning, arrows.meta}
\usepackage{blindtext}
\usepackage{todonotes}
\usepackage{enumitem}
\usepackage{pifont}
\usepackage{multirow}
\usepackage{tabularx}
\usepackage{xspace}

\DeclareMathOperator*{\argmax}{argmax}

\aclfinalcopy %

\newcommand{\cqa}{cQA\xspace}

\title{Neural Duplicate Question Detection without Labeled Training Data}

\author{Andreas R\"uckl\'e
\and
Nafise Sadat Moosavi
\and
Iryna Gurevych\\
	Ubiquitous Knowledge Processing Lab (UKP)\\
	Department of Computer Science, Technische Universit\"{a}t Darmstadt\\
	{\url{www.ukp.tu-darmstadt.de}}\\
}

\date{}

\begin{document}
\maketitle
\begin{abstract}
    Supervised training of neural models to duplicate question detection in community Question Answering (\cqa) requires large amounts of labeled question pairs, 
    which are costly to obtain.
    To minimize this cost, recent works thus often used alternative methods, e.g., adversarial domain adaptation. %
    In this work, we propose two novel methods: (1) the automatic generation of duplicate questions, and (2) weak supervision using the title and body of a question. We show that both can achieve improved performances even though they do not require any labeled data. We provide comprehensive comparisons of popular training strategies,
    which provides important insights on how to best train models in different scenarios. %
    We show that our proposed approaches are more effective in many cases because they can utilize larger amounts of unlabeled data from \cqa forums.
    Finally, we also show that our proposed approach for weak supervision with question title and body information is also an effective method to train \cqa answer selection models without direct answer supervision.

\end{abstract}

\section{Introduction}
\nocite{rueckle:AAAI:2019}

The automatic detection of question duplicates in community Question Answering (\cqa) forums is an important task that can help users to more effectively find existing questions and answers \cite{nakov2017semeval,Cao2012,Xue:2008,Jeon2005}, and to avoid posting similar questions multiple times.
Neural approaches to duplicate detection typically require large quantities of labeled question pairs for supervised training---i.e., labeled pairs of duplicate questions that can be answered with the same information.\footnote{For example, ``Nautilus shortcut for new blank files?'' and ``How do you create a new document keyboard shortcut?'' are titles of labeled duplicate questions from AskUbuntu.com.}

In practice, it is often difficult to obtain such data because of the immense manual effort that is required for annotation. 
A large number of \cqa forums thus do not contain enough labeled data for supervised training of neural models.\footnote{\citet{Shah2018} argue that even larger StackExchange sites do not offer enough duplicates for supervised training. %
Further, there exist many platforms that do not contain any labeled duplicates (e.g., \url{https://gutefrage.net}).}
Therefore, 
recent works have used 
alternative training methods.
This includes weak supervision with question-answer pairs~\cite{Qiu2015}, semi-supervised %
training ~\cite{Uva2018}, 
and adversarial domain transfer~\cite{Shah2018}. 
An important 
limitation of these methods is that they still rely on substantial amounts of labeled data---either thousands of duplicate questions (e.g., from a similar source domain in the case of domain transfer) or large numbers of question-answer pairs. 
Furthermore, 
unsupervised methods
rely on encoder-decoder architectures that impose limitations on the model architectures and they often fall short of the performances that are achieved with supervised training \cite{Lei2016}, or they need to be combined with complex features to achieve state-of-the-art results \cite{Zhang2018}.
To train effective duplicate question detection models for the large number of \cqa forums without labeled duplicates we thus
need other methods %
that do not require any annotations %
while performing
on-par with supervised in-domain training.

In this work, we 
propose two novel methods %
for
scenarios where we only have access to unlabeled questions (title-body)%
, including
(1) automatic duplicate question generation (\emph{DQG}); and (2) weak supervision with the title-body pairs (\emph{WS-TB}). 
Because a question body typically provides additional important information that is not included in the title \cite{wu-etal-2018-question}, 
we %
hypothesize that titles and bodies have similar properties as duplicate questions. %
For instance, 
they are only partially redundant but fundamentally describe the same question
(see Figure~\ref{fig:introduction:example} for an example). 
As a consequence, we can use the information from titles and bodies together with their relations to train our models. %

In DQG, we use question generation models to generate a new question title from the body and then consider the generated title as a duplicate to the question's original title. %
In WS-TB, 
we take this one step further and directly train models on title-body pairs---i.e., learning to predict whether both texts %
belong to the same question.
The advantage of our proposed methods, compared to previous work, is that 
 they can make use of the large number of unlabeled questions (titles and bodies) in \cqa forums, %
 which is typically an order of magnitude more data than is available for supervised training.%
\footnote{
Question titles and bodies are common in all StackExchange sites, popular platforms in other languages (e.g., GuteFrage.net), and forums such as Reddit. A counterexample is Quora, which only contains question titles. However, there exists a large annotated corpus of question pairs for this forum.
}

In our experiments, we evaluate common question retrieval and duplicate detection models such as RCNN \cite{Lei2016} and BiLSTM %
and compare a wide range of training methods: DQG, WS-TB, supervised training, adversarial domain transfer, weak supervision with question-answer pairs, and unsupervised training.
We perform extensive experiments on multiple datasets %
and compare the different training methods 
in different scenarios, 
which provides important insights on how to best train models %
with varying amounts training data.
We show that:
\begin{enumerate}
    \item Training models with title-body information is very effective. %
    With larger
    amounts of unlabeled questions, WS-TB and DQG outperform adversarial domain transfer from similar source domains by more than 5.8pp on average. Because the amounts of labeled question duplicates is often limited, WS-TB and DQG can in some cases achieve better performances than supervised training. %
    \item DQG transfers well across domains, i.e., question generation models can be applied to novel target domains to obtain generated duplicates that are suitable for model training. %
    \item Our training methods are effective when being used to fine-tune more recent models such as BERT~\cite{Devlin2018}.
    \item WS-TB can also be used to train \cqa answer selection models without direct answer supervision. %
    This shows that our methods can have broader impact on related tasks and beyond duplicate question detection.

\end{enumerate}
\begin{figure}

\centering

\begin{tikzpicture}[xscale=.77,yscale=.77]

\definecolor{lightgreen}{RGB}{205,245,195}
\definecolor{lightblue}{RGB}{228,232,252}
\definecolor{lightyellow}{RGB}{255,255,220}
\definecolor{lightgray}{RGB}{245,245,245}

\node(titleheader)[draw=none,align=left,anchor=north west,font=\scriptsize] at (0,8) 
{\textbf{TITLE}};

\node(title)[draw,below=0.5mm of titleheader.south west,align=left,anchor=north west,text width=7.4cm,font=\footnotesize,fill=lightgreen] 
{How to customize each Firefox window icon individually?};

\node(bodyheader)[draw=none,below=1.5mm of title.south west,align=left,anchor=north west,font=\scriptsize]
{\textbf{BODY (1\textsuperscript{st} PARAGRAPH)}};

\node(body)[draw,below=0.5mm of bodyheader.south west,align=left,anchor=north west,text width=7.4cm,font=\footnotesize,fill=lightblue] 
{I'm a tab hoarder and I admit it. But at least I've sorted them into contextual windows now, and I'd love to have different icons for each window in the Windows task bar (not the tab bar, which is governed by the favicons). How can this be achieved?};

\node(answersheader)[draw=none,below=2mm of body.south west,align=left,anchor=north west,font=\scriptsize]
{\textbf{ANSWER}};
\node(answers)[draw,below=0.5mm of answersheader.south west,align=left,anchor=north west,text width=7.4cm,font=\footnotesize,fill=lightgray] 
{This can be done using the free AutoHotkey. Create a .ahk text file and enter these contents:  (~$\ldots$~)};

\draw[-] ([xshift=5mm]title.south) to[out=270,in=90] ([xshift=5mm]body.north);

\end{tikzpicture}
\caption{An example question, the first paragraph of its body, and the first answer (from SuperUser\footnotemark). %
}
\label{fig:introduction:example}
\end{figure}

\footnotetext{\url{https://superuser.com/q/1393090}}  
\section{Related Work}

Duplicate question detection is closely related to question-question similarity and question retrieval. 
Early approaches %
use translation models \cite{Jeon2005,Xue:2008,Zhou2011} that were further enhanced with question category information \cite{Cao2012} and topic models \cite{Ji2012,Zhang2014}. 

More recent works in the context of the SemEval \cqa challenges \cite{nakov2017semeval} improve upon this and use tree kernels (TK) \cite{Martino2016}, TK with neural networks \cite{Romeo2016},
neural networks with multi-task learning \cite{Bonadiman2017}, and encoder-decoder architectures together with shallow lexical matching and mismatching \cite{Zhang2018}.
Common neural models such as CNNs achieved superior performance compared to TK when they were trained on sufficiently large numbers of labeled question pairs \cite{Uva2018}.

Similarly, neural representation learning methods have proved to be most effective in technical \cqa domains. \citet{DosSantos2015}, for example, learn representations of questions with CNNs and compare them with cosine similarity for scoring. \citet{Lei2016} %
 propose RCNN, which extends CNN with a recurrent mechanism (adaptive gated decay). 
This approach was further extended with question-type information \cite{Gupta2018}.

If in-domain training data is scarce---i.e., if the \cqa platform does not offer enough labeled duplicates---%
alternative training strategies are required.
If there exist some labeled question pairs (thousands), one can first train a less data-hungry non-neural model and use 
it for supervised training of neural models
\cite{Uva2018}. 
Further, if there exist large numbers of labeled question-answer pairs, we can use them for weakly-supervised training \cite{Wang2017-cnn,Qiu2015}.

More related to our work are methods that do not rely on any labeled data in the target domain. %
Existing methods use unsupervised training with encoder-decoder architectures \cite{Lei2016,Zhang2018},
and adversarial domain transfer where the model is trained on a source domain and adversarially adapted to a target domain %
\cite{Shah2018}.
However, such approaches typically fall short of the performances that are being achieved with in-domain supervised training. %

In contrast, we propose two novel methods, DQG and WS-TB, that do not require any annotations for model training and in some cases perform better than in-domain supervised training with duplicate questions. 
While WS-TB is related to the approaches mentioned before, DQG is is also related to question generation (QG). 
Most of the previous work in QG is in the context of reading comprehension \citep[e.g.,][]{Du2017:ACL,Subramanian2018,Zhao2018,Du2018:ACL} or QG for question answering \cite{Duan2017}. 
They substantially differ from our approach because they 
generate questions based on specific answer spans, while DQG generates
a new title from a question's body that can be used as a question duplicate. %
\section{Training Methods}

\begin{table}
  \centering
  \small
  \begin{tabular}{l|ccc}
    \toprule
    \textbf{Method} & \textbf{Duplicates} & \textbf{Answers} & \textbf{Bodies} \\
    \midrule
    Supervised & \ding{53} & - & \textcolor{gray}{(\ding{53})} \\
    WS-QA & - & \ding{53} & \textcolor{gray}{(\ding{53})} \\
    Domain Transfer & ~~\ding{53}$^*$ & - & \textcolor{gray}{(\ding{53})} \\
    \midrule
    DQG & - & - & \ding{53} \\
    WS-TB & - & - & \ding{53} \\
    \bottomrule
  \end{tabular}
  \caption{The different training methods and the data they use. Models typically also use text from the bodies during training and evaluation, which we indicate with \textcolor{gray}{(\ding{53})}. \ding{53}$^*$~=~domain transfer requires duplicates from a sufficiently similar source domain.}
  \label{tbl:methods:overview}
\end{table} 
Given a pair of questions, our goal is to determine whether they are duplicates or not. 
In practice, the model predictions are often used to rank a list of potentially similar questions in regard to a new user question, e.g., to retrieve the most likely duplicate for automatic question answering.
To train models, we obtain a set of examples
$\{(x_1,y_1), \ldots, (x_N,y_N)\}$ 
in which each ${x_n \in \mathcal{X}}$ is an instance, i.e., a tuple containing texts such as two questions, and ${y_n \in \{-1, +1\}}$ is its corresponding binary label, e.g., duplicate/no-duplicate.
Obtaining instances with positive labels $\mathcal{X}^+ = \left\{ x_n^+ \in \mathcal{X} | y_n = 1 \right\}$ is generally more difficult than obtaining $\mathcal{X}^-$ because instances with negative labels can be automatically generated (e.g., by randomly sampling unrelated questions). %

In the following, we outline three existing training methods that use different kinds of instances, and in \S\ref{sec:training:ours} we present our two novel methods: duplicate question generation, and weak supervision with title-body pairs. 
Both do not require any annotations in $\mathcal{X}^+$, and can therefore use larger amounts of data from the \cqa forums.
Table~\ref{tbl:methods:overview} gives an overview of the different training methods. %

\subsection{Existing Methods}
\label{sec:training:existing}

\paragraph{Supervised (in-domain) training} is the most common method, %
which requires labeled question duplicates, i.e.,
${x_n^+ = \left(q_n, \tilde{q}_n\right)}$. 
Unrelated questions can be randomly sampled. %
With this data, we can train representation learning models \cite[e.g.,][]{Lei2016} or pairwise classifiers \cite[e.g.,][]{Uva2018}.
Most models combine the titles and bodies of the questions during training and evaluation (e.g., by concatenation), which can improve performances \cite{Lei2016,wu-etal-2018-question}. %

\paragraph{Weak supervision with question-answer pairs (WS-QA)}
is an alternative to supervised training for larger platforms without duplicate annotations \cite{Qiu2015}. %
WS-QA trains models with questions $q_n$ and accepted answers $a_n$, and therefore ${x_n^+ = \left(q_n, a_n\right)}$.
Instances in $X^-$ can be obtained by randomly sampling unrelated answers for a question.
An advantage 
of this method 
is that there typically exist more labeled answers than duplicate questions. For instance, Yahoo! answers has accepted answers but it does not contain labeled duplicate questions. %

\paragraph{Domain transfer}
performs supervised training in a source domain and applies the trained model to a different target domain in which no labeled duplicate questions exist. %
\citet{Shah2018} use this method with adversarial training 
to learn domain-invariant question representations prior to transfer.
They show that adversarial training can considerably improve upon direct transfer, but their method requires sufficiently similar source and target domains. For instance, they could not successfully transfer models between technical and 
other non-technical  domains.

\subsection{Proposed Methods with Unlabeled Data} %
\label{sec:training:ours}

The disadvantage of the existing methods is that they require labeled question duplicates, accepted answers, or similar source and target domains for transfer.
We could alternatively use unsupervised training within an encoder-decoder framework, %
but this imposes important limitations on the network architecture, e.g., a question can only be encoded independently (no inter-attention). %

Our proposed methods 
do not suffer from these drawbacks, i.e., they do not require labeled data and they do not impose architectural limitations. %

\begin{figure}

\centering

\begin{tikzpicture}[xscale=.77,yscale=.77]

\definecolor{lightgreen}{RGB}{205,245,195}
\definecolor{lightblue}{RGB}{228,232,252}
\definecolor{lightyellow}{RGB}{255,255,220}
\definecolor{lightgray}{RGB}{245,245,245}

\tikzset{
    titlestyle/.style = {
        fill=lightgreen,
        text width=1.5cm,
        text centered,
        font=\footnotesize,
        draw
    }
}
\tikzset{
    paragraphstyle/.style = {
        fill=lightgray,
        text width=1.5cm,
        minimum height=0.8cm,
        text centered,
        font=\footnotesize,
        draw
    }
}
\tikzset{
    paragraphstyle_select/.style = {
        fill=lightblue,
        text width=1.5cm,
        minimum height=0.8cm,
        text centered,
        font=\footnotesize,
        draw
    }
}
\tikzset{
    paragraphstyle_noselect/.style = {
        fill=white,
        text width=1.5cm,
        minimum height=0.8cm,
        text centered,
        font=\footnotesize,
        text=gray,
        draw=gray
    }
}
\tikzset{
    bodystyle/.style = {
        fill=lightgray,
        text width=1.5cm,
        minimum height=1.75cm,
        text centered,
        font=\footnotesize,
        draw
    }
}

\node(title1)[titlestyle,anchor=north west] at (0,0) 
{Title};
\node(body1)[paragraphstyle_select,anchor=north west,below=1.5mm of title1.south] 
{Body};
\draw[-] (title1.south) to[out=270,in=90] (body1.north);

\node(trainlabel)[draw=none,above left=7mm and 1mm of title1.north west,align=left,anchor=north west,font=\scriptsize]
{\textbf{TRAINING}};

\node(title2)[titlestyle,anchor=north west,right=1.8cm of title1.east]
{Title};
\node(body2)[paragraphstyle_select,anchor=north west,below=1.5mm of title2.south] 
{Body};
\draw[-] (title2.south) to[out=270,in=90] (body2.north);

\node(title3)[titlestyle,anchor=north west,right=6mm of title2.east]
{Duplicate};

\node(inferencelabel)[draw=none,above left=7mm and 2mm of title2.north west,align=left,anchor=north west,font=\scriptsize]
{\textbf{DATA GENERATION}};

\draw[-{Latex[scale=1.1]}] (body1.east) to[out=0,in=0,looseness=1.4] (title1.east)
node[right=6mm,rotate=270,font=\footnotesize] {restore};

\draw[-{Latex[scale=1.1]}] (body2.east) to[out=0,in=270,looseness=1] (title3.south)
node[below=6mm,rotate=0,font=\footnotesize] {generate};

\draw[densely dotted] (title2.east) to[out=0,in=180,looseness=0] (title3.west);

\end{tikzpicture}
\caption{During training we restore the original question title from its body. During data generation we consider the generated title as a new duplicate question.}
\label{fig:train-inference}
\end{figure}
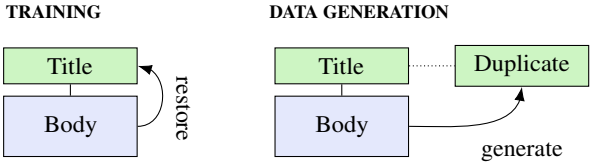

\paragraph{Duplicate question generation (DQG)}
generates new question titles from question bodies, which we then consider as duplicates to the original titles.
Our overall approach is depicted in Figure~\ref{fig:train-inference}.
First, we train a question generation model $\mathrm{QG}$ to %
maximize $P(\mathrm{title}(q_n)|\mathrm{body}(q_n))$.
This is similar to news headline generation or abstractive summarization \cite{Rush2015,Chopra2016} because $\mathrm{QG}$ 
needs to identify the most relevant aspects in the body that best characterize the question.
However, restoring the exact title is usually not possible 
because titles and bodies often contain complementary information (see, e.g., Figure~\ref{fig:introduction:example}). %
We therefore %
consider ${\mathrm{QG}(\mathrm{body}(q_n))}$ as a duplicate of $\mathrm{title}(q_n)$ and obtain positive labeled instances 
${x_n^+ = \left(\mathrm{title}(q_n), \mathrm{QG}(\mathrm{body}(q_n))\right)}$.

Because DQG requires no annotated data, we can use this method to train duplicate detection models for all \cqa forums that offer a reasonable number of unlabeled title-body pairs to obtain a suitable $\mathrm{QG}$ model (the smallest number of questions we tried for training of question generation models is 23k, see \S\ref{sec:additional:domain-transfer}). %
An important advantage is that
we can make use of \emph{all questions} %
(after some basic filtering),
which is often an order of magnitude more training data than annotated duplicates.

We can use any sequence to sequence model for $\mathrm{QG}$, and we performed experiments with a Transformer \cite{Vaswani2017} and MQAN \cite{McCann2018decaNLP}.

\paragraph{Weak supervision with title-body pairs (WS-TB)}
takes the assumption of DQG one step further.
If question titles and question bodies have similar attributes as duplicates, we could also just train duplicate detection models directly on this data without prior question generation.
In WS-TB, we thus %
train models to predict whether a given title and body are related, i.e., whether they belong to the same question. %
Therefore, 
${x^+ = \left(\mathrm{title}(q_n), \mathrm{body}(q_n)\right)}$.

This method considerably simplifies the sourcing of training data because it requires no separate question generation model. However, it also means that the duplicate detection model must be able to handle texts of considerably different lengths during training (for instance, bodies in SuperUser.com have an average length of 125 words). This might not be suitable for some text matching models, e.g., ones that were designed to compare two sentences. %
\section{Experiments}

\subsection{Experimental Setup}
\label{sec:experiments:setup}

\begin{table}
    \centering
    \footnotesize
    \begin{tabular}{lrrr}
        \toprule
         \textbf{Dataset} & \textbf{Train / Dev / Test} & $|$\textbf{Q}$|$ & $|$\textbf{A}$|$ \\
        \midrule
        AskUbuntu-Lei & 12,584 / ~~200 / ~~200 & 288k & 84k \\
        AskUbuntu & 9106 / 1000 / 1000 & 288k & 84k \\
        SuperUser & 9106 / 1000 / 1000 & 377k & 142k \\
        Apple & - / 1000 / 1000 & 89k & 29k \\
        Android & - / 1000 / 1000 & 47k & 14k \\
        \bottomrule
    \end{tabular}
    \caption{The dataset statistics. Numbers for Train/Dev/Test refer to the number of questions with duplicates. %
    $|$Q$|$ refers to the number of unlabeled questions, and $|$A$|$ refers to the number of accepted answers.}
    \label{tbl:setup:data}
\end{table}
 
We use models and data from previous literature to obtain comparable results for evaluation, and we rely on their official implementations, default hyperparameters, and evaluation measures.
An overview of the datasets is given in Table \ref{tbl:setup:data}, which also shows that they considerably differ in the amounts of data that is available for the different training methods.

The evaluation setup is the same for all datasets: given a user question $q$ and a list of potentially related questions, the goal is to re-rank this list to retrieve duplicates of $q$ (one or more potential related questions are labeled as duplicates). 
Even though some training methods do not use bodies during training, e.g., DQG, during evaluation they use the same data (annotated pairs of questions with titles and bodies).\footnote{It has been shown that including bodies in the experimental setup can lead to improved performances \cite{Lei2016}. In initial experiments, we found that the performances are mostly impacted by having access to bodies during evaluation.}

\paragraph{AskUbuntu-Lei.}
First, we replicate the setup of \citet{Lei2016}, which uses RCNN to learn dense vector representations of questions and then compares them with cosine similarity for scoring. Besides supervised training, this also includes unsupervised training with the encoder-decoder architecture.
We report precision at 5 (P@5), i.e., how many of the top-5 ranked questions are %
actual
duplicates. The dataset is based on the AskUbuntu data of \citet{DosSantos2015} with additional manual annotations for dev/test splits (user questions have an average of 5.7 duplicates). %

\paragraph{Android, Apple, AskUbuntu, and SuperUser.}
Second, we replicate the setup of \citet{Shah2018}, which %
uses BiLSTM to learn question representations. This setup also includes adversarial domain transfer. 
The data is from the AskUbuntu, SuperUser, Android, and Apple sites of StackExchange, and different to AskUbuntu-Lei, each question has only one duplicate. 
We measure AUC(0.05), which is the area under curve with a threshold for false positives---\citet{Shah2018} argue that this is more stable when there are many unrelated questions.

\paragraph{Questions and answers.}
To train the models with WS-TB and WS-QA, we use 
questions and answers from publicly available data dumps\footnote{\url{https://archive.org/download/stackexchange}} 
of the StackExchange platforms. %
We obtain our new training sets as specified in \S\ref{sec:training:ours}. For instance, for WS-TB we replace every annotated duplicate ${(q_n, \tilde q_n)}$ from the original training split with ${(\mathrm{title}(q_n), \mathrm{body}(q_n))}$, and we randomly sample unrelated bodies to obtain training instances with negative labels.

It is important to note that the number of questions and answers is much larger than the number of annotated duplicate questions. %
Therefore, we can add more instances to the training splits with these methods. However, if not otherwise noted, we use the same number of training instances as in the original training splits with duplicates. %

\paragraph{DQG setup.}
To train question generation models, we use the same StackExchange data.
We filter the questions 
to ensure that the bodies contain multiple sentences. 
Further, if a body contains multiple paragraphs, we only keep the one with the highest similarity to the title. Details of the filtering approach are included in the Appendix. Less than 10\% of the questions are discarded on average. 

We train a MQAN (Multi-task Question Answering Network) model, which was proposed as a very general network architecture to solve a wide variety of tasks as part of the Natural Language Decathlon \cite{McCann2018decaNLP}.
The model first encodes the input with LSTMs and applies different attention mechanisms,
including multi-headed self-attention. 
MQAN also includes pointer-generator networks \cite{See2017}, which allow it to copy tokens from the input text depending on the attention distribution of an earlier layer.

We performed the same experiments with a Transformer sequence to sequence model \cite{Vaswani2017}, but on average MQAN performed better because of its ability to copy words and phrases from the body.
We include the Transformer results and a comparison with MQAN in the Appendix.

We use all available questions from a \cqa forum to train the question generation model. We perform early stopping using BLEU scores to avoid overfitting. %
To generate duplicate questions, we then apply the trained model on all questions from the same \cqa forum. We do not use a separate heldout set %
because this would considerably limit both the question generation training data and the number of generated duplicates. %
We did not observe negative effects from using this procedure. %

\subsection{Experimental Results}
\label{sec:results}

\begin{table*}
    \centering
    \footnotesize
    \begin{tabular}{l|r|rrrrr}
        \toprule
         & \textbf{AskUbuntu-Lei} & \textbf{Android} & \textbf{Apple} & \textbf{AskUbuntu} & \textbf{SuperUser} & Average \\
        \multicolumn{2}{r}{Measuring P@5. Results (dev / test) for RCNN} & \multicolumn{5}{|c}{Measuring AUC(0.05). Results for BiLSTM} \\
        \midrule
        \multicolumn{6}{l}{\textbf{Trained on 1x data}~~\scriptsize{(all methods use the same number of training instances as in supervised training)}} \\
        \midrule
        Supervised (in-domain) & \underline{\textbf{48.0}} / 45.0 & - & - & \textbf{0.848} & \textbf{0.944} & - \\
        Unsupervised & 42.6 / 42.0 & - & - & - & - & - \\
        Direct Transfer (best) & - & 0.770 & 0.828 & 0.730 & 0.908 & 0.809 \\
        Adversarial Transfer (best) & - & 0.790 & 0.861 & 0.796 & 0.911 & 0.840 \\
        WS-QA & 47.2 / 45.3 & 0.780 & \textbf{0.894} & 0.790 & 0.919 & 0.846 \\
        DQG & 46.4 / 44.8 & 0.793 & 0.870 & 0.801 & 0.921 & 0.846 \\
        WS-TB & 46.4 / \underline{\textbf{45.4}} & \textbf{0.811} & 0.866 & 0.804 & 0.913 & \textbf{0.849} \\
        \midrule
        \multicolumn{6}{l}{\textbf{Trained on all available data}} \\
        \midrule
        Unsupervised & 43.0 / 41.8 & - & - & - & - & - \\
        WS-QA & 47.3 / 44.2 & 0.814 & 0.901 & 0.828 & 0.951 & 0.874 \\
        DQG & \textbf{47.4} / 44.3 & 0.833 & 0.911 & 0.855 & 0.944 & 0.886 \\
        WS-TB & 47.3 / \textbf{45.3} & 0.852 & 0.910 & \underline{\textbf{0.871}} & \underline{\textbf{0.952}} & 0.896 \\
        DQG + WS-TB (combined) & 46.4 / 44.0 & \underline{\textbf{0.863}} & \underline{\textbf{0.916}} & 0.866 & 0.946 & \underline{\textbf{0.898}} \\
        \bottomrule
    \end{tabular}
    \caption{Results of the models with different training strategies. Android and Apple datasets do not contain labeled duplicates for supervised in-domain training.}
    \label{tbl:results:main}
\end{table*} The results are given in Table \ref{tbl:results:main}. %
For domain transfer, we report the best scores from \citet{Shah2018}, which reflects an optimal transfer setup from a similar source domain.

\paragraph{Supervised training.}

As we expect, supervised in-domain training with labeled duplicates achieves better scores compared to other training methods when we consider the same number of training instances.
An exception is on AskUbuntu-Lei where DQG, WS-TB, and WS-QA can achieve results that are on the same level on test or marginally worse on dev. 

One reason for the better performances with labeled duplicates is that they contain more information, i.e., a pair of questions consist of two titles and two bodies compared to just one title and body for each training instance in WS-TB.
However, the results show that all weakly supervised techniques as well as DQG are 
effective training methods. %

\paragraph{DQG, WS-TB, and WS-QA.}

All methods outperform direct transfer from a similar source domain as well as the encoder-decoder approach on AskUbuntu-Lei. 
On average, WS-TB is the most effective method, and it consistently outperforms adversarial domain transfer (0.9pp on average). 

We otherwise do not observe large differences between the three methods DQG, WS-TB, and WS-QA, which shows that (1) the models we use can learn from different text lengths (title-body, question-answer); and (2) the information that we extract in DQG is suitable for training (examples are given in \S\ref{sec:analysis}).
The good results of WS-TB might suggest that question generation as separate step is not required, however we argue that it can be important in a number of scenarios, e.g., when we need to train sentence matching models that would otherwise not be able to handle long texts.

\paragraph{Using all available data.}
One of the biggest advantages of our proposed methods is that they can use larger amounts of training data. This greatly improves the model performances for BiLSTM, where we observe average improvements of up to 4.7pp (for WS-TB). In many cases our methods now perform better than supervised training. We observe smaller improvements for WS-QA (2.8pp on avg) because it has access to fewer training instances. The performances for RCNN on AskUbuntu-Lei are mostly unchanged with minor improvements on dev. The reason is that the performances were already close to supervised training with the same data sizes. %

\begin{figure}
    \centering
    \footnotesize
\begin{tikzpicture}[x=1pt,y=1pt]
\definecolor{fillColor}{RGB}{255,255,255}
\path[use as bounding box,fill=fillColor,fill opacity=0.00] (0,0) rectangle (216.81,202.36);
\begin{scope}
\path[clip] (  0.00,  0.00) rectangle (216.81,202.36);
\definecolor{drawColor}{RGB}{255,255,255}
\definecolor{fillColor}{RGB}{255,255,255}

\path[draw=drawColor,line width= 0.6pt,line join=round,line cap=round,fill=fillColor] ( -0.00,  0.00) rectangle (216.81,202.36);
\end{scope}
\begin{scope}
\path[clip] ( 32.35,135.83) rectangle (111.45,185.55);
\definecolor{fillColor}{RGB}{255,255,255}

\path[fill=fillColor] ( 32.35,135.83) rectangle (111.45,185.55);
\definecolor{drawColor}{gray}{0.92}

\path[draw=drawColor,line width= 0.6pt,line join=round] ( 32.35,143.60) --
	(111.45,143.60);

\path[draw=drawColor,line width= 0.6pt,line join=round] ( 32.35,154.62) --
	(111.45,154.62);

\path[draw=drawColor,line width= 0.6pt,line join=round] ( 32.35,165.65) --
	(111.45,165.65);

\path[draw=drawColor,line width= 0.6pt,line join=round] ( 32.35,176.68) --
	(111.45,176.68);

\path[draw=drawColor,line width= 0.6pt,line join=round] ( 37.80,135.83) --
	( 37.80,185.55);

\path[draw=drawColor,line width= 0.6pt,line join=round] ( 60.53,135.83) --
	( 60.53,185.55);

\path[draw=drawColor,line width= 0.6pt,line join=round] ( 83.26,135.83) --
	( 83.26,185.55);

\path[draw=drawColor,line width= 0.6pt,line join=round] (105.99,135.83) --
	(105.99,185.55);
\definecolor{drawColor}{RGB}{0,114,178}

\path[draw=drawColor,line width= 1.1pt,line join=round] ( 37.80,138.09) --
	( 60.53,138.09) --
	( 83.26,138.09) --
	(105.99,138.09);
\definecolor{drawColor}{RGB}{86,180,233}

\path[draw=drawColor,line width= 1.1pt,line join=round] ( 37.80,149.11) --
	( 60.53,149.11) --
	( 83.26,149.11) --
	(105.99,149.11);
\definecolor{drawColor}{RGB}{0,158,115}

\path[draw=drawColor,line width= 1.1pt,line join=round] ( 37.80,160.69) --
	( 60.53,164.55) --
	( 83.26,178.33) --
	(105.99,183.29);
\definecolor{fillColor}{RGB}{0,158,115}

\path[draw=drawColor,line width= 0.4pt,line join=round,line cap=round,fill=fillColor] ( 37.80,160.69) circle (  1.96);

\path[draw=drawColor,line width= 0.4pt,line join=round,line cap=round,fill=fillColor] ( 60.53,164.55) circle (  1.96);

\path[draw=drawColor,line width= 0.4pt,line join=round,line cap=round,fill=fillColor] ( 83.26,178.33) circle (  1.96);

\path[draw=drawColor,line width= 0.4pt,line join=round,line cap=round,fill=fillColor] (105.99,183.29) circle (  1.96);
\definecolor{drawColor}{gray}{0.20}

\path[draw=drawColor,line width= 0.6pt,line join=round,line cap=round] ( 32.35,135.83) rectangle (111.45,185.55);
\end{scope}
\begin{scope}
\path[clip] ( 32.35, 50.77) rectangle (111.45,100.50);
\definecolor{fillColor}{RGB}{255,255,255}

\path[fill=fillColor] ( 32.35, 50.77) rectangle (111.45,100.50);
\definecolor{drawColor}{gray}{0.92}

\path[draw=drawColor,line width= 0.6pt,line join=round] ( 32.35, 55.09) --
	(111.45, 55.09);

\path[draw=drawColor,line width= 0.6pt,line join=round] ( 32.35, 65.36) --
	(111.45, 65.36);

\path[draw=drawColor,line width= 0.6pt,line join=round] ( 32.35, 75.63) --
	(111.45, 75.63);

\path[draw=drawColor,line width= 0.6pt,line join=round] ( 32.35, 85.91) --
	(111.45, 85.91);

\path[draw=drawColor,line width= 0.6pt,line join=round] ( 32.35, 96.18) --
	(111.45, 96.18);

\path[draw=drawColor,line width= 0.6pt,line join=round] ( 37.80, 50.77) --
	( 37.80,100.50);

\path[draw=drawColor,line width= 0.6pt,line join=round] ( 51.44, 50.77) --
	( 51.44,100.50);

\path[draw=drawColor,line width= 0.6pt,line join=round] ( 65.08, 50.77) --
	( 65.08,100.50);

\path[draw=drawColor,line width= 0.6pt,line join=round] ( 78.72, 50.77) --
	( 78.72,100.50);

\path[draw=drawColor,line width= 0.6pt,line join=round] ( 92.35, 50.77) --
	( 92.35,100.50);

\path[draw=drawColor,line width= 0.6pt,line join=round] (105.99, 50.77) --
	(105.99,100.50);
\definecolor{drawColor}{RGB}{0,114,178}

\path[draw=drawColor,line width= 1.1pt,line join=round] ( 37.80, 53.03) --
	( 51.44, 53.03) --
	( 65.08, 53.03) --
	( 78.72, 53.03) --
	( 92.35, 53.03) --
	(105.99, 53.03);
\definecolor{drawColor}{RGB}{86,180,233}

\path[draw=drawColor,line width= 1.1pt,line join=round] ( 37.80, 56.11) --
	( 51.44, 56.11) --
	( 65.08, 56.11) --
	( 78.72, 56.11) --
	( 92.35, 56.11) --
	(105.99, 56.11);
\definecolor{drawColor}{RGB}{230,159,0}

\path[draw=drawColor,line width= 1.1pt,line join=round] ( 37.80, 90.51) --
	( 51.44, 90.51) --
	( 65.08, 90.51) --
	( 78.72, 90.51) --
	( 92.35, 90.51) --
	(105.99, 90.51);
\definecolor{drawColor}{RGB}{0,158,115}

\path[draw=drawColor,line width= 1.1pt,line join=round] ( 37.80, 58.17) --
	( 51.44, 66.39) --
	( 65.08, 72.55) --
	( 78.72, 82.83) --
	( 92.35, 92.07) --
	(105.99, 98.24);
\definecolor{fillColor}{RGB}{0,158,115}

\path[draw=drawColor,line width= 0.4pt,line join=round,line cap=round,fill=fillColor] ( 37.80, 58.17) circle (  1.96);

\path[draw=drawColor,line width= 0.4pt,line join=round,line cap=round,fill=fillColor] ( 51.44, 66.39) circle (  1.96);

\path[draw=drawColor,line width= 0.4pt,line join=round,line cap=round,fill=fillColor] ( 65.08, 72.55) circle (  1.96);

\path[draw=drawColor,line width= 0.4pt,line join=round,line cap=round,fill=fillColor] ( 78.72, 82.83) circle (  1.96);

\path[draw=drawColor,line width= 0.4pt,line join=round,line cap=round,fill=fillColor] ( 92.35, 92.07) circle (  1.96);

\path[draw=drawColor,line width= 0.4pt,line join=round,line cap=round,fill=fillColor] (105.99, 98.24) circle (  1.96);
\definecolor{drawColor}{gray}{0.20}

\path[draw=drawColor,line width= 0.6pt,line join=round,line cap=round] ( 32.35, 50.77) rectangle (111.45,100.50);
\end{scope}
\begin{scope}
\path[clip] (137.71,135.83) rectangle (216.81,185.55);
\definecolor{fillColor}{RGB}{255,255,255}

\path[fill=fillColor] (137.71,135.83) rectangle (216.81,185.55);
\definecolor{drawColor}{gray}{0.92}

\path[draw=drawColor,line width= 0.6pt,line join=round] (137.71,144.70) --
	(216.81,144.70);

\path[draw=drawColor,line width= 0.6pt,line join=round] (137.71,155.73) --
	(216.81,155.73);

\path[draw=drawColor,line width= 0.6pt,line join=round] (137.71,166.75) --
	(216.81,166.75);

\path[draw=drawColor,line width= 0.6pt,line join=round] (137.71,177.78) --
	(216.81,177.78);

\path[draw=drawColor,line width= 0.6pt,line join=round] (143.17,135.83) --
	(143.17,185.55);

\path[draw=drawColor,line width= 0.6pt,line join=round] (160.22,135.83) --
	(160.22,185.55);

\path[draw=drawColor,line width= 0.6pt,line join=round] (177.26,135.83) --
	(177.26,185.55);

\path[draw=drawColor,line width= 0.6pt,line join=round] (194.31,135.83) --
	(194.31,185.55);

\path[draw=drawColor,line width= 0.6pt,line join=round] (211.36,135.83) --
	(211.36,185.55);
\definecolor{drawColor}{RGB}{0,114,178}

\path[draw=drawColor,line width= 1.1pt,line join=round] (143.17,138.09) --
	(160.22,138.09) --
	(177.26,138.09) --
	(194.31,138.09) --
	(211.36,138.09);
\definecolor{drawColor}{RGB}{86,180,233}

\path[draw=drawColor,line width= 1.1pt,line join=round] (143.17,156.28) --
	(160.22,156.28) --
	(177.26,156.28) --
	(194.31,156.28) --
	(211.36,156.28);
\definecolor{drawColor}{RGB}{0,158,115}

\path[draw=drawColor,line width= 1.1pt,line join=round] (143.17,159.03) --
	(160.22,166.75) --
	(177.26,170.06) --
	(194.31,182.19) --
	(211.36,183.29);
\definecolor{fillColor}{RGB}{0,158,115}

\path[draw=drawColor,line width= 0.4pt,line join=round,line cap=round,fill=fillColor] (143.17,159.03) circle (  1.96);

\path[draw=drawColor,line width= 0.4pt,line join=round,line cap=round,fill=fillColor] (160.22,166.75) circle (  1.96);

\path[draw=drawColor,line width= 0.4pt,line join=round,line cap=round,fill=fillColor] (177.26,170.06) circle (  1.96);

\path[draw=drawColor,line width= 0.4pt,line join=round,line cap=round,fill=fillColor] (194.31,182.19) circle (  1.96);

\path[draw=drawColor,line width= 0.4pt,line join=round,line cap=round,fill=fillColor] (211.36,183.29) circle (  1.96);
\definecolor{drawColor}{gray}{0.20}

\path[draw=drawColor,line width= 0.6pt,line join=round,line cap=round] (137.71,135.83) rectangle (216.81,185.55);
\end{scope}
\begin{scope}
\path[clip] (137.71, 50.77) rectangle (216.81,100.50);
\definecolor{fillColor}{RGB}{255,255,255}

\path[fill=fillColor] (137.71, 50.77) rectangle (216.81,100.50);
\definecolor{drawColor}{gray}{0.92}

\path[draw=drawColor,line width= 0.6pt,line join=round] (137.71, 62.65) --
	(216.81, 62.65);

\path[draw=drawColor,line width= 0.6pt,line join=round] (137.71, 75.47) --
	(216.81, 75.47);

\path[draw=drawColor,line width= 0.6pt,line join=round] (137.71, 88.30) --
	(216.81, 88.30);

\path[draw=drawColor,line width= 0.6pt,line join=round] (143.17, 50.77) --
	(143.17,100.50);

\path[draw=drawColor,line width= 0.6pt,line join=round] (156.81, 50.77) --
	(156.81,100.50);

\path[draw=drawColor,line width= 0.6pt,line join=round] (170.44, 50.77) --
	(170.44,100.50);

\path[draw=drawColor,line width= 0.6pt,line join=round] (184.08, 50.77) --
	(184.08,100.50);

\path[draw=drawColor,line width= 0.6pt,line join=round] (197.72, 50.77) --
	(197.72,100.50);

\path[draw=drawColor,line width= 0.6pt,line join=round] (211.36, 50.77) --
	(211.36,100.50);
\definecolor{drawColor}{RGB}{0,114,178}

\path[draw=drawColor,line width= 1.1pt,line join=round] (143.17, 53.03) --
	(156.81, 53.03) --
	(170.44, 53.03) --
	(184.08, 53.03) --
	(197.72, 53.03) --
	(211.36, 53.03);
\definecolor{drawColor}{RGB}{86,180,233}

\path[draw=drawColor,line width= 1.1pt,line join=round] (143.17, 74.19) --
	(156.81, 74.19) --
	(170.44, 74.19) --
	(184.08, 74.19) --
	(197.72, 74.19) --
	(211.36, 74.19);
\definecolor{drawColor}{RGB}{230,159,0}

\path[draw=drawColor,line width= 1.1pt,line join=round] (143.17, 90.86) --
	(156.81, 90.86) --
	(170.44, 90.86) --
	(184.08, 90.86) --
	(197.72, 90.86) --
	(211.36, 90.86);
\definecolor{drawColor}{RGB}{0,158,115}

\path[draw=drawColor,line width= 1.1pt,line join=round] (143.17, 76.76) --
	(156.81, 84.45) --
	(170.44, 87.66) --
	(184.08, 92.79) --
	(197.72, 95.99) --
	(211.36, 98.24);
\definecolor{fillColor}{RGB}{0,158,115}

\path[draw=drawColor,line width= 0.4pt,line join=round,line cap=round,fill=fillColor] (143.17, 76.76) circle (  1.96);

\path[draw=drawColor,line width= 0.4pt,line join=round,line cap=round,fill=fillColor] (156.81, 84.45) circle (  1.96);

\path[draw=drawColor,line width= 0.4pt,line join=round,line cap=round,fill=fillColor] (170.44, 87.66) circle (  1.96);

\path[draw=drawColor,line width= 0.4pt,line join=round,line cap=round,fill=fillColor] (184.08, 92.79) circle (  1.96);

\path[draw=drawColor,line width= 0.4pt,line join=round,line cap=round,fill=fillColor] (197.72, 95.99) circle (  1.96);

\path[draw=drawColor,line width= 0.4pt,line join=round,line cap=round,fill=fillColor] (211.36, 98.24) circle (  1.96);
\definecolor{drawColor}{gray}{0.20}

\path[draw=drawColor,line width= 0.6pt,line join=round,line cap=round] (137.71, 50.77) rectangle (216.81,100.50);
\end{scope}
\begin{scope}
\path[clip] ( 32.35,100.50) rectangle (111.45,117.30);
\definecolor{drawColor}{gray}{0.20}
\definecolor{fillColor}{gray}{0.85}

\path[draw=drawColor,line width= 0.6pt,line join=round,line cap=round,fill=fillColor] ( 32.35,100.50) rectangle (111.45,117.30);
\definecolor{drawColor}{gray}{0.10}

\node[text=drawColor,anchor=base,inner sep=0pt, outer sep=0pt, scale=  0.88] at ( 71.90,105.87) {SuperUser};
\end{scope}
\begin{scope}
\path[clip] (137.71,100.50) rectangle (216.81,117.30);
\definecolor{drawColor}{gray}{0.20}
\definecolor{fillColor}{gray}{0.85}

\path[draw=drawColor,line width= 0.6pt,line join=round,line cap=round,fill=fillColor] (137.71,100.50) rectangle (216.81,117.30);
\definecolor{drawColor}{gray}{0.10}

\node[text=drawColor,anchor=base,inner sep=0pt, outer sep=0pt, scale=  0.88] at (177.26,105.87) {AskUbuntu};
\end{scope}
\begin{scope}
\path[clip] ( 32.35,185.55) rectangle (111.45,202.36);
\definecolor{drawColor}{gray}{0.20}
\definecolor{fillColor}{gray}{0.85}

\path[draw=drawColor,line width= 0.6pt,line join=round,line cap=round,fill=fillColor] ( 32.35,185.55) rectangle (111.45,202.36);
\definecolor{drawColor}{gray}{0.10}

\node[text=drawColor,anchor=base,inner sep=0pt, outer sep=0pt, scale=  0.88] at ( 71.90,190.92) {Android};
\end{scope}
\begin{scope}
\path[clip] (137.71,185.55) rectangle (216.81,202.36);
\definecolor{drawColor}{gray}{0.20}
\definecolor{fillColor}{gray}{0.85}

\path[draw=drawColor,line width= 0.6pt,line join=round,line cap=round,fill=fillColor] (137.71,185.55) rectangle (216.81,202.36);
\definecolor{drawColor}{gray}{0.10}

\node[text=drawColor,anchor=base,inner sep=0pt, outer sep=0pt, scale=  0.88] at (177.26,190.92) {Apple};
\end{scope}
\begin{scope}
\path[clip] (  0.00,  0.00) rectangle (216.81,202.36);
\definecolor{drawColor}{gray}{0.20}

\path[draw=drawColor,line width= 0.6pt,line join=round] ( 37.80, 48.02) --
	( 37.80, 50.77);

\path[draw=drawColor,line width= 0.6pt,line join=round] ( 51.44, 48.02) --
	( 51.44, 50.77);

\path[draw=drawColor,line width= 0.6pt,line join=round] ( 65.08, 48.02) --
	( 65.08, 50.77);

\path[draw=drawColor,line width= 0.6pt,line join=round] ( 78.72, 48.02) --
	( 78.72, 50.77);

\path[draw=drawColor,line width= 0.6pt,line join=round] ( 92.35, 48.02) --
	( 92.35, 50.77);

\path[draw=drawColor,line width= 0.6pt,line join=round] (105.99, 48.02) --
	(105.99, 50.77);
\end{scope}
\begin{scope}
\path[clip] (  0.00,  0.00) rectangle (216.81,202.36);
\definecolor{drawColor}{RGB}{0,0,0}

\node[text=drawColor,anchor=base,inner sep=0pt, outer sep=0pt, scale=  0.89] at ( 37.80, 39.69) {1x};

\node[text=drawColor,anchor=base,inner sep=0pt, outer sep=0pt, scale=  0.89] at ( 51.44, 39.69) {2x};

\node[text=drawColor,anchor=base,inner sep=0pt, outer sep=0pt, scale=  0.89] at ( 65.08, 39.69) {4x};

\node[text=drawColor,anchor=base,inner sep=0pt, outer sep=0pt, scale=  0.89] at ( 78.72, 39.69) {8x};

\node[text=drawColor,anchor=base,inner sep=0pt, outer sep=0pt, scale=  0.89] at ( 92.35, 39.69) {16x};

\node[text=drawColor,anchor=base,inner sep=0pt, outer sep=0pt, scale=  0.89] at (105.99, 39.69) {All};
\end{scope}
\begin{scope}
\path[clip] (  0.00,  0.00) rectangle (216.81,202.36);
\definecolor{drawColor}{gray}{0.20}

\path[draw=drawColor,line width= 0.6pt,line join=round] (143.17, 48.02) --
	(143.17, 50.77);

\path[draw=drawColor,line width= 0.6pt,line join=round] (156.81, 48.02) --
	(156.81, 50.77);

\path[draw=drawColor,line width= 0.6pt,line join=round] (170.44, 48.02) --
	(170.44, 50.77);

\path[draw=drawColor,line width= 0.6pt,line join=round] (184.08, 48.02) --
	(184.08, 50.77);

\path[draw=drawColor,line width= 0.6pt,line join=round] (197.72, 48.02) --
	(197.72, 50.77);

\path[draw=drawColor,line width= 0.6pt,line join=round] (211.36, 48.02) --
	(211.36, 50.77);
\end{scope}
\begin{scope}
\path[clip] (  0.00,  0.00) rectangle (216.81,202.36);
\definecolor{drawColor}{RGB}{0,0,0}

\node[text=drawColor,anchor=base,inner sep=0pt, outer sep=0pt, scale=  0.89] at (143.17, 39.69) {1x};

\node[text=drawColor,anchor=base,inner sep=0pt, outer sep=0pt, scale=  0.89] at (156.81, 39.69) {2x};

\node[text=drawColor,anchor=base,inner sep=0pt, outer sep=0pt, scale=  0.89] at (170.44, 39.69) {4x};

\node[text=drawColor,anchor=base,inner sep=0pt, outer sep=0pt, scale=  0.89] at (184.08, 39.69) {8x};

\node[text=drawColor,anchor=base,inner sep=0pt, outer sep=0pt, scale=  0.89] at (197.72, 39.69) {16x};

\node[text=drawColor,anchor=base,inner sep=0pt, outer sep=0pt, scale=  0.89] at (211.36, 39.69) {All};
\end{scope}
\begin{scope}
\path[clip] (  0.00,  0.00) rectangle (216.81,202.36);
\definecolor{drawColor}{gray}{0.20}

\path[draw=drawColor,line width= 0.6pt,line join=round] ( 37.80,133.08) --
	( 37.80,135.83);

\path[draw=drawColor,line width= 0.6pt,line join=round] ( 60.53,133.08) --
	( 60.53,135.83);

\path[draw=drawColor,line width= 0.6pt,line join=round] ( 83.26,133.08) --
	( 83.26,135.83);

\path[draw=drawColor,line width= 0.6pt,line join=round] (105.99,133.08) --
	(105.99,135.83);
\end{scope}
\begin{scope}
\path[clip] (  0.00,  0.00) rectangle (216.81,202.36);
\definecolor{drawColor}{RGB}{0,0,0}

\node[text=drawColor,anchor=base,inner sep=0pt, outer sep=0pt, scale=  0.89] at ( 37.80,124.75) {1x};

\node[text=drawColor,anchor=base,inner sep=0pt, outer sep=0pt, scale=  0.89] at ( 60.53,124.75) {2x};

\node[text=drawColor,anchor=base,inner sep=0pt, outer sep=0pt, scale=  0.89] at ( 83.26,124.75) {4x};

\node[text=drawColor,anchor=base,inner sep=0pt, outer sep=0pt, scale=  0.89] at (105.99,124.75) {All};
\end{scope}
\begin{scope}
\path[clip] (  0.00,  0.00) rectangle (216.81,202.36);
\definecolor{drawColor}{gray}{0.20}

\path[draw=drawColor,line width= 0.6pt,line join=round] (143.17,133.08) --
	(143.17,135.83);

\path[draw=drawColor,line width= 0.6pt,line join=round] (160.22,133.08) --
	(160.22,135.83);

\path[draw=drawColor,line width= 0.6pt,line join=round] (177.26,133.08) --
	(177.26,135.83);

\path[draw=drawColor,line width= 0.6pt,line join=round] (194.31,133.08) --
	(194.31,135.83);

\path[draw=drawColor,line width= 0.6pt,line join=round] (211.36,133.08) --
	(211.36,135.83);
\end{scope}
\begin{scope}
\path[clip] (  0.00,  0.00) rectangle (216.81,202.36);
\definecolor{drawColor}{RGB}{0,0,0}

\node[text=drawColor,anchor=base,inner sep=0pt, outer sep=0pt, scale=  0.89] at (143.17,124.75) {1x};

\node[text=drawColor,anchor=base,inner sep=0pt, outer sep=0pt, scale=  0.89] at (160.22,124.75) {2x};

\node[text=drawColor,anchor=base,inner sep=0pt, outer sep=0pt, scale=  0.89] at (177.26,124.75) {4x};

\node[text=drawColor,anchor=base,inner sep=0pt, outer sep=0pt, scale=  0.89] at (194.31,124.75) {8x};

\node[text=drawColor,anchor=base,inner sep=0pt, outer sep=0pt, scale=  0.89] at (211.36,124.75) {All};
\end{scope}
\begin{scope}
\path[clip] (  0.00,  0.00) rectangle (216.81,202.36);
\definecolor{drawColor}{RGB}{0,0,0}

\node[text=drawColor,anchor=base east,inner sep=0pt, outer sep=0pt, scale=  0.89] at (132.76,141.64) {0.84};

\node[text=drawColor,anchor=base east,inner sep=0pt, outer sep=0pt, scale=  0.89] at (132.76,152.66) {0.86};

\node[text=drawColor,anchor=base east,inner sep=0pt, outer sep=0pt, scale=  0.89] at (132.76,163.69) {0.88};

\node[text=drawColor,anchor=base east,inner sep=0pt, outer sep=0pt, scale=  0.89] at (132.76,174.71) {0.90};
\end{scope}
\begin{scope}
\path[clip] (  0.00,  0.00) rectangle (216.81,202.36);
\definecolor{drawColor}{gray}{0.20}

\path[draw=drawColor,line width= 0.6pt,line join=round] (134.96,144.70) --
	(137.71,144.70);

\path[draw=drawColor,line width= 0.6pt,line join=round] (134.96,155.73) --
	(137.71,155.73);

\path[draw=drawColor,line width= 0.6pt,line join=round] (134.96,166.75) --
	(137.71,166.75);

\path[draw=drawColor,line width= 0.6pt,line join=round] (134.96,177.78) --
	(137.71,177.78);
\end{scope}
\begin{scope}
\path[clip] (  0.00,  0.00) rectangle (216.81,202.36);
\definecolor{drawColor}{RGB}{0,0,0}

\node[text=drawColor,anchor=base east,inner sep=0pt, outer sep=0pt, scale=  0.89] at (132.76, 59.59) {0.76};

\node[text=drawColor,anchor=base east,inner sep=0pt, outer sep=0pt, scale=  0.89] at (132.76, 72.41) {0.80};

\node[text=drawColor,anchor=base east,inner sep=0pt, outer sep=0pt, scale=  0.89] at (132.76, 85.23) {0.84};
\end{scope}
\begin{scope}
\path[clip] (  0.00,  0.00) rectangle (216.81,202.36);
\definecolor{drawColor}{gray}{0.20}

\path[draw=drawColor,line width= 0.6pt,line join=round] (134.96, 62.65) --
	(137.71, 62.65);

\path[draw=drawColor,line width= 0.6pt,line join=round] (134.96, 75.47) --
	(137.71, 75.47);

\path[draw=drawColor,line width= 0.6pt,line join=round] (134.96, 88.30) --
	(137.71, 88.30);
\end{scope}
\begin{scope}
\path[clip] (  0.00,  0.00) rectangle (216.81,202.36);
\definecolor{drawColor}{RGB}{0,0,0}

\node[text=drawColor,anchor=base east,inner sep=0pt, outer sep=0pt, scale=  0.89] at ( 27.40,140.53) {0.78};

\node[text=drawColor,anchor=base east,inner sep=0pt, outer sep=0pt, scale=  0.89] at ( 27.40,151.56) {0.80};

\node[text=drawColor,anchor=base east,inner sep=0pt, outer sep=0pt, scale=  0.89] at ( 27.40,162.59) {0.82};

\node[text=drawColor,anchor=base east,inner sep=0pt, outer sep=0pt, scale=  0.89] at ( 27.40,173.61) {0.84};
\end{scope}
\begin{scope}
\path[clip] (  0.00,  0.00) rectangle (216.81,202.36);
\definecolor{drawColor}{gray}{0.20}

\path[draw=drawColor,line width= 0.6pt,line join=round] ( 29.60,143.60) --
	( 32.35,143.60);

\path[draw=drawColor,line width= 0.6pt,line join=round] ( 29.60,154.62) --
	( 32.35,154.62);

\path[draw=drawColor,line width= 0.6pt,line join=round] ( 29.60,165.65) --
	( 32.35,165.65);

\path[draw=drawColor,line width= 0.6pt,line join=round] ( 29.60,176.68) --
	( 32.35,176.68);
\end{scope}
\begin{scope}
\path[clip] (  0.00,  0.00) rectangle (216.81,202.36);
\definecolor{drawColor}{RGB}{0,0,0}

\node[text=drawColor,anchor=base east,inner sep=0pt, outer sep=0pt, scale=  0.89] at ( 27.40, 52.02) {0.91};

\node[text=drawColor,anchor=base east,inner sep=0pt, outer sep=0pt, scale=  0.89] at ( 27.40, 62.30) {0.92};

\node[text=drawColor,anchor=base east,inner sep=0pt, outer sep=0pt, scale=  0.89] at ( 27.40, 72.57) {0.93};

\node[text=drawColor,anchor=base east,inner sep=0pt, outer sep=0pt, scale=  0.89] at ( 27.40, 82.84) {0.94};

\node[text=drawColor,anchor=base east,inner sep=0pt, outer sep=0pt, scale=  0.89] at ( 27.40, 93.12) {0.95};
\end{scope}
\begin{scope}
\path[clip] (  0.00,  0.00) rectangle (216.81,202.36);
\definecolor{drawColor}{gray}{0.20}

\path[draw=drawColor,line width= 0.6pt,line join=round] ( 29.60, 55.09) --
	( 32.35, 55.09);

\path[draw=drawColor,line width= 0.6pt,line join=round] ( 29.60, 65.36) --
	( 32.35, 65.36);

\path[draw=drawColor,line width= 0.6pt,line join=round] ( 29.60, 75.63) --
	( 32.35, 75.63);

\path[draw=drawColor,line width= 0.6pt,line join=round] ( 29.60, 85.91) --
	( 32.35, 85.91);

\path[draw=drawColor,line width= 0.6pt,line join=round] ( 29.60, 96.18) --
	( 32.35, 96.18);
\end{scope}
\begin{scope}
\path[clip] (  0.00,  0.00) rectangle (216.81,202.36);
\definecolor{drawColor}{RGB}{0,0,0}

\node[text=drawColor,anchor=base,inner sep=0pt, outer sep=0pt, scale=  1.00] at (124.58, 28.11) {Training Data};
\end{scope}
\begin{scope}
\path[clip] (  0.00,  0.00) rectangle (216.81,202.36);
\definecolor{drawColor}{RGB}{0,0,0}

\node[text=drawColor,rotate= 90.00,anchor=base,inner sep=0pt, outer sep=0pt, scale=  1.00] at (  6.89,118.16) {AUC~(0.05)};
\end{scope}
\begin{scope}

\definecolor{drawColor}{RGB}{0,0,0}

\path[draw=drawColor,line width=0.4pt] ( 29.68,  0.00) rectangle (219.48, 20.48);
\end{scope}
\begin{scope}
\path[clip] (  0.00,  0.00) rectangle (216.81,202.36);
\definecolor{fillColor}{RGB}{255,255,255}

\end{scope}
\begin{scope}
\path[clip] (  0.00,  0.00) rectangle (216.81,202.36);
\definecolor{fillColor}{RGB}{255,255,255}

\end{scope}
\begin{scope}
\path[clip] (  0.00,  0.00) rectangle (216.81,202.36);
\definecolor{drawColor}{RGB}{0,114,178}

\path[draw=drawColor,line width= 1.1pt,line join=round] ( 32.70, 15.36) -- ( 40.89, 15.36);
\end{scope}
\begin{scope}
\path[clip] (  0.00,  0.00) rectangle (216.81,202.36);
\definecolor{fillColor}{RGB}{255,255,255}

\end{scope}
\begin{scope}
\path[clip] (  0.00,  0.00) rectangle (216.81,202.36);
\definecolor{drawColor}{RGB}{86,180,233}

\path[draw=drawColor,line width= 1.1pt,line join=round] ( 32.70,  5.12) -- ( 40.89,  5.12);
\end{scope}
\begin{scope}
\path[clip] (  0.00,  0.00) rectangle (216.81,202.36);
\definecolor{fillColor}{RGB}{255,255,255}

\end{scope}
\begin{scope}
\path[clip] (  0.00,  0.00) rectangle (216.81,202.36);
\definecolor{drawColor}{RGB}{230,159,0}

\path[draw=drawColor,line width= 1.1pt,line join=round] (125.10, 15.36) -- (133.29, 15.36);
\end{scope}
\begin{scope}
\path[clip] (  0.00,  0.00) rectangle (216.81,202.36);
\definecolor{fillColor}{RGB}{255,255,255}

\end{scope}
\begin{scope}
\path[clip] (  0.00,  0.00) rectangle (216.81,202.36);
\definecolor{drawColor}{RGB}{0,158,115}

\path[draw=drawColor,line width= 1.1pt,line join=round] (125.10,  5.12) -- (133.29,  5.12);
\end{scope}
\begin{scope}
\path[clip] (  0.00,  0.00) rectangle (216.81,202.36);
\definecolor{fillColor}{RGB}{0,158,115}

\path[fill=fillColor] (129.20,  5.12) circle (  1.96);
\end{scope}
\begin{scope}
\path[clip] (  0.00,  0.00) rectangle (216.81,202.36);
\definecolor{drawColor}{RGB}{0,0,0}

\node[text=drawColor,anchor=base west,inner sep=0pt, outer sep=0pt, scale=  0.88] at ( 41.92, 12.33) {Direct transfer (best)};
\end{scope}
\begin{scope}
\path[clip] (  0.00,  0.00) rectangle (216.81,202.36);
\definecolor{drawColor}{RGB}{0,0,0}

\node[text=drawColor,anchor=base west,inner sep=0pt, outer sep=0pt, scale=  0.88] at ( 41.92,  2.09) {Adv. transfer (best)};
\end{scope}
\begin{scope}
\path[clip] (  0.00,  0.00) rectangle (216.81,202.36);
\definecolor{drawColor}{RGB}{0,0,0}

\node[text=drawColor,anchor=base west,inner sep=0pt, outer sep=0pt, scale=  0.88] at (134.32, 12.33) {Supervised (in-domain)};
\end{scope}
\begin{scope}
\path[clip] (  0.00,  0.00) rectangle (216.81,202.36);
\definecolor{drawColor}{RGB}{0,0,0}

\node[text=drawColor,anchor=base west,inner sep=0pt, outer sep=0pt, scale=  0.88] at (134.32,  2.09) {WS-TB};
\end{scope}
\end{tikzpicture}
     \caption{Performances of BiLSTM as a function of the available training data. %
    `2x'
    means that 
    there are twice as many (unlabeled) questions available to WS-TB than there are annotated duplicate questions in the original dataset
    (1x = 9106).}
    \label{fig:results:shahplot}
\end{figure}
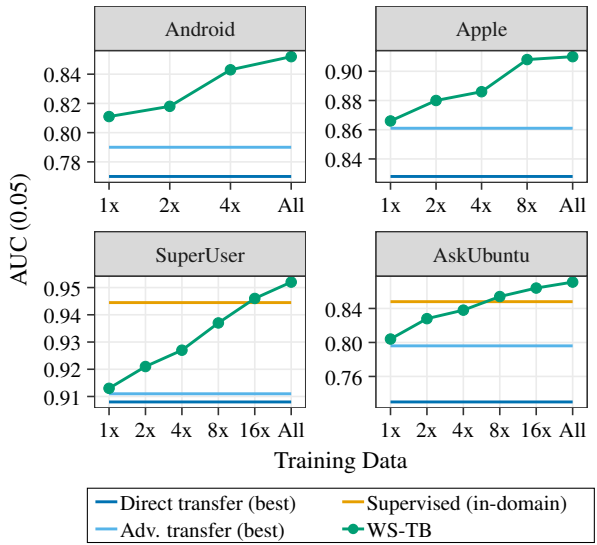

In Figure \ref{fig:results:shahplot} we 
plot the performance scores of BiLSTM averaged over the four StackExchange datasets in relation to the available training data with WS-TB.
We see that the model performance consistently improves when we increase the training data
(we observe similar trends for DQG and WS-QA). 
Thus, it is crucial to make use of all available data from the \cqa forums. %
We also explored a combination of our two proposed approaches where we merge their respective training sets. 
We find that this helps mostly for smaller \cqa platforms with fewer questions (where larger training sets would be most necessary), e.g., the performances on Android and Apple improve by 0.6--1.1pp compared to WS-TB.
Even though the combination does not introduce new information because both use the same question data, complementing WS-TB with DQG can provide additional variation %
with the generative component.

In summary, our results show that even when we have access to sufficient numbers of labeled duplicates, the best method is not always supervised training. When we use larger numbers of title-body pairs, DQG and WS-TB can achieve better performances. %
\section{Further Application Scenarios}
\label{sec:additional:domain-transfer}
To test if our methods are applicable to other scenarios with high practical relevance,
we explore
(1) whether DQG can be used in \cqa forums with fewer unlabeled title-body pairs, (2) if we can use WS-TB to train answer selection models without labeled question-answer pair, and (3) how well large pre-trained language models perform when being fine-tuned with our methods.

\subsection{DQG for Small-Scale \cqa Forums}
In our previous experiments, we assumed that there exist enough unlabeled questions %
to train the question generation model (at least 47k questions, see Table \ref{tbl:setup:data}).
To simulate a more challenging scenario with fewer in-domain questions, we explore the effects of cross-domain question generation.
This is highly relevant for DQG because %
in such scenarios the generated duplicates could be combined with WS-TB to obtain more training data. %

We replicate the transfer setup of \citet{Shah2018} where they originally transfer the duplicate question detection model from a source to a target domain. For DQG we instead train the question generation model on the source domain and generate duplicates for the target domain, with which we then train the duplicate detection model. %
To provide a fair comparison against adversarial domain transfer, we always use the same number of 9106 duplicates to train the duplicate detection models. %

Results for the transfer from SuperUser and AskUbuntu to other domains are given in Table \ref{tbl:results:transfer}.
They show that the question generation model for DQG can be successfully transferred across similar domains with only minor effects on the 
performances. Importantly, DQG still performs better than adversarial domain transfer with the same number of training instances. %

To test an even more extreme case, we also transfer from StackExchange Academia (only 23k title-body pairs to train question generation) 
to the technical target domains. This could, e.g., be realistic for other languages with fewer \cqa forums. %
Most notably, the performance of DQG decreases only mildly, which demonstrates its practical applicability in even more challenging scenarios. This is mostly due to the copy mechanism of MQAN, which is stable across domains (see \S\ref{sec:analysis}).
\begin{table}
    \centering
    \footnotesize
    \begin{tabular}{ll|r|rr}
        \toprule
        \textbf{Source} & \textbf{Target} & \textbf{Adv. DT} & \textbf{DQG} & $\Delta$ \\
        \midrule
        \multirow{3}{*}{AskUbuntu} 
         & Android & 0.790 & \textbf{0.797} & $+$0.004\\
         & Apple & 0.855 & \textbf{0.861} & $-$0.009\\
         & SuperUser & 0.911 & \textbf{0.916} & $-$0.005 \\
         \midrule
        \multirow{3}{*}{SuperUser} 
         & Android & 0.790 & \textbf{0.794} & $+$0.001 \\
         & Apple & \textbf{0.861} & \textbf{0.861} & $-$0.009 \\
         & AskUbuntu & 0.796 & \textbf{0.809} & $+$0.008 \\
         \midrule
         \multirow{3}{*}{Academia} 
         & Android & - & 0.776 & $-$0.017 \\
         & Apple & - & 0.854 & $-$0.016 \\
         & SuperUser & - & 0.912 & $-$0.009 \\
         & AskUbuntu & - & 0.760 & $-$0.041 \\
         \bottomrule
    \end{tabular}
    \caption{The domain transfer performances. $\Delta$ denotes the difference to the setup with in-domain DQG.}
    \label{tbl:results:transfer}
\end{table} 
\subsection{Answer Selection}
\label{sec:additional:answer-selection}

In answer selection we predict 
whether a candidate answer is relevant in regard to a question
 \cite{Tay2017,nakov2017semeval,Tan2016,rueckle:2017:IWCS}, which is similar to duplicate question detection.

To test whether our strategy to train models with title-body pairs is also suitable for answer selection, we use the data and code of \citet{rueckle:AAAI:2019} and train two different types of models with WS-TB on their five datasets that are based on StackExchange Apple, Aviation, Academia, Cooking, and Travel. We train (1) a siamese BiLSTM, which learns question and answer representations; and (2) their neural relevance matching model COALA. 
Both are evaluated by how well they re-rank a list of candidate answers in regard to a question.

The results are given in Table \ref{tbl:results:coala} where we report the accuracy (P@1), averaged over the five datasets. 
Interestingly, we do not observe large differences between supervised training and WS-TB for both models when they use the same number of positive training instances (ranging from 2.8k to 5.8k). Thus, using title-body information instead of question-answer pairs to train models without direct answer supervision is feasible and effective.
Further, when we use all available title-body pairs, the BiLSTM model substantially improves by 5pp, which is only slightly worse than COALA (which was designed for smaller training sets). 
We hypothesize that one reason is that BiLSTM can learn improved representations with the additional data. Further, title-body pairs have a higher overlap than question-answer pairs (see \S\ref{sec:analysis}) which provides a stronger training signal to the siamese network. %

These results demonstrate that our work can have broader impact to \cqa%
, e.g., to train models on other tasks beyond duplicate question detection. %

\begin{table}
    \centering
    \footnotesize
    \begin{tabular}{lrrr}
        \toprule
        \textbf{Model} & \textbf{Supervised} & \textbf{WS-TB} (1x) & \textbf{WS-TB} (all) \\
        \midrule
        BiLSTM & 35.3 & 37.5 & 42.5\\
        COALA & 44.7 & 45.2 & 44.5 \\
        \bottomrule
    \end{tabular}
    \caption{Answer selection performances (averaged over five datasets) when trained with question-answer pairs vs. WS-TB.}
    \label{tbl:results:coala}
\end{table} 
\subsection{BERT Fine-Tuning}
\label{sec:additional:bert}

Large pre-trained language models such as BERT \cite{devlin2018bert} and RoBERTa \cite{roberta} have recently led to considerable improvements across a wide range of NLP tasks.
To test whether our training strategies can also be used to fine-tune such models, we integrate BERT in the setups of our previous experiments.\footnote{We add the AskUbuntu-Lei dataset to the framework of \citet{rueckle:AAAI:2019} for our BERT experiments. Details are given in the Appendix.} %
We fine-tune a pre-trained BERT-base (uncased) model with supervised training, WS-TB (1x), and WS-TB (8x). %

The results are given in Table \ref{tbl:results:bert}. 
We observe similar trends as before but with overall better results. When increasing the number of training examples, the model performances consistently improve.
We note that we have also conducted preliminary experiments with larger BERT models where we observed further improvements. %

\begin{table*}
    \centering
    \footnotesize
    \begin{tabular}{l|r|rrrr|r}
        \toprule
         & \textbf{AskUbuntu-Lei} & \textbf{Android} & \textbf{Apple} & \textbf{AskUbuntu} & \textbf{SuperUser} & \textbf{Answer Selection} \\
        \multicolumn{2}{r}{Measuring P@5. Results for dev / test} & \multicolumn{4}{|c|}{Measuring AUC(0.05)} & Measuring Accuracy\\
        \midrule
        Supervised (in-domain) & \textbf{54.0} / \textbf{52.3} & - & - & 0.862 & 0.954 & 56.8 \\
        WS-TB (1x) & 47.8 / 47.2 & 0.857 & 0.908 & 0.841 & 0.932 & 55.5 \\
        WS-TB (8x) & 50.4 / 49.6 & \textbf{0.896} & \textbf{0.933} & \textbf{0.897} & \textbf{0.971} & \textbf{59.7} \\
        \bottomrule
    \end{tabular}
    \caption{Results of BERT fine-tuning with different training strategies. The answer selection accuracies are averaged over the five StackExchange datasets of \citep{rueckle:AAAI:2019} (individual scores are in Appendix~\ref{ap:bert-details}).}
    \label{tbl:results:bert}
\end{table*} 

\section{Analysis}
\label{sec:analysis}

\subsection{Overlap}

To analyze the differences 
in the training methods we calculate the overlap between the texts of positive training instances (e.g., question-question, title-body, question-answer etc.). For questions, we concatenate titles and bodies.

\begin{figure}
    \centering
    \footnotesize
\begin{tikzpicture}[x=1pt,y=1pt]
\definecolor{fillColor}{RGB}{255,255,255}
\path[use as bounding box,fill=fillColor,fill opacity=0.00] (0,0) rectangle (187.90,108.41);
\begin{scope}
\path[clip] (  0.00,  0.00) rectangle (187.90,108.41);
\definecolor{drawColor}{RGB}{255,255,255}
\definecolor{fillColor}{RGB}{255,255,255}

\path[draw=drawColor,line width= 0.6pt,line join=round,line cap=round,fill=fillColor] (  0.00,  0.00) rectangle (187.90,108.41);
\end{scope}
\begin{scope}
\path[clip] ( 16.32, 39.86) rectangle (187.90,108.41);
\definecolor{fillColor}{RGB}{255,255,255}

\path[fill=fillColor] ( 16.32, 39.86) rectangle (187.90,108.41);
\definecolor{drawColor}{gray}{0.92}

\path[draw=drawColor,line width= 0.6pt,line join=round] ( 16.32, 39.86) --
	(187.90, 39.86);

\path[draw=drawColor,line width= 0.6pt,line join=round] ( 16.32, 54.14) --
	(187.90, 54.14);

\path[draw=drawColor,line width= 0.6pt,line join=round] ( 16.32, 68.42) --
	(187.90, 68.42);

\path[draw=drawColor,line width= 0.6pt,line join=round] ( 16.32, 82.70) --
	(187.90, 82.70);

\path[draw=drawColor,line width= 0.6pt,line join=round] ( 16.32, 96.98) --
	(187.90, 96.98);
\definecolor{fillColor}{RGB}{86,180,233}

\path[fill=fillColor] ( 32.92, 39.86) rectangle ( 42.44, 61.28);
\definecolor{fillColor}{RGB}{0,114,178}

\path[fill=fillColor] ( 23.41, 39.86) rectangle ( 32.92, 92.70);
\definecolor{fillColor}{RGB}{86,180,233}

\path[fill=fillColor] ( 67.52, 39.86) rectangle ( 77.03, 54.14);
\definecolor{fillColor}{RGB}{0,114,178}

\path[fill=fillColor] ( 58.00, 39.86) rectangle ( 67.52, 74.13);
\definecolor{fillColor}{RGB}{86,180,233}

\path[fill=fillColor] (102.11, 39.86) rectangle (111.62, 48.43);
\definecolor{fillColor}{RGB}{0,114,178}

\path[fill=fillColor] ( 92.60, 39.86) rectangle (102.11, 66.99);
\definecolor{fillColor}{RGB}{86,180,233}

\path[fill=fillColor] (136.70, 39.86) rectangle (146.22, 55.57);
\definecolor{fillColor}{RGB}{0,114,178}

\path[fill=fillColor] (127.19, 39.86) rectangle (136.70, 94.12);
\definecolor{fillColor}{RGB}{86,180,233}

\path[fill=fillColor] (171.30, 39.86) rectangle (180.81, 84.13);
\definecolor{fillColor}{RGB}{0,114,178}

\path[fill=fillColor] (161.78, 39.86) rectangle (171.30,102.69);
\definecolor{drawColor}{gray}{0.20}

\path[draw=drawColor,line width= 0.6pt,line join=round,line cap=round] ( 16.32, 39.86) rectangle (187.90,108.41);
\end{scope}
\begin{scope}
\path[clip] (  0.00,  0.00) rectangle (187.90,108.41);
\definecolor{drawColor}{RGB}{0,0,0}

\node[text=drawColor,anchor=base east,inner sep=0pt, outer sep=0pt, scale=  0.89] at ( 11.37, 36.79) {0.0};

\node[text=drawColor,anchor=base east,inner sep=0pt, outer sep=0pt, scale=  0.89] at ( 11.37, 51.07) {0.1};

\node[text=drawColor,anchor=base east,inner sep=0pt, outer sep=0pt, scale=  0.89] at ( 11.37, 65.35) {0.2};

\node[text=drawColor,anchor=base east,inner sep=0pt, outer sep=0pt, scale=  0.89] at ( 11.37, 79.63) {0.3};

\node[text=drawColor,anchor=base east,inner sep=0pt, outer sep=0pt, scale=  0.89] at ( 11.37, 93.92) {0.4};
\end{scope}
\begin{scope}
\path[clip] (  0.00,  0.00) rectangle (187.90,108.41);
\definecolor{drawColor}{gray}{0.20}

\path[draw=drawColor,line width= 0.6pt,line join=round] ( 13.57, 39.86) --
	( 16.32, 39.86);

\path[draw=drawColor,line width= 0.6pt,line join=round] ( 13.57, 54.14) --
	( 16.32, 54.14);

\path[draw=drawColor,line width= 0.6pt,line join=round] ( 13.57, 68.42) --
	( 16.32, 68.42);

\path[draw=drawColor,line width= 0.6pt,line join=round] ( 13.57, 82.70) --
	( 16.32, 82.70);

\path[draw=drawColor,line width= 0.6pt,line join=round] ( 13.57, 96.98) --
	( 16.32, 96.98);
\end{scope}
\begin{scope}
\path[clip] (  0.00,  0.00) rectangle (187.90,108.41);
\definecolor{drawColor}{gray}{0.20}

\path[draw=drawColor,line width= 0.6pt,line join=round] ( 32.92, 37.11) --
	( 32.92, 39.86);

\path[draw=drawColor,line width= 0.6pt,line join=round] ( 67.52, 37.11) --
	( 67.52, 39.86);

\path[draw=drawColor,line width= 0.6pt,line join=round] (102.11, 37.11) --
	(102.11, 39.86);

\path[draw=drawColor,line width= 0.6pt,line join=round] (136.70, 37.11) --
	(136.70, 39.86);

\path[draw=drawColor,line width= 0.6pt,line join=round] (171.30, 37.11) --
	(171.30, 39.86);
\end{scope}
\begin{scope}
\path[clip] (  0.00,  0.00) rectangle (187.90,108.41);
\definecolor{drawColor}{RGB}{0,0,0}

\node[text=drawColor,anchor=base,inner sep=0pt, outer sep=0pt, scale=  0.89] at ( 32.92, 28.78) {Supervised};

\node[text=drawColor,anchor=base,inner sep=0pt, outer sep=0pt, scale=  0.89] at ( 32.92, 19.17) {(duplicates)};

\node[text=drawColor,anchor=base,inner sep=0pt, outer sep=0pt, scale=  0.89] at ( 67.52, 28.78) {WS-QA};

\node[text=drawColor,anchor=base,inner sep=0pt, outer sep=0pt, scale=  0.89] at (102.11, 28.78) {WS-QA};

\node[text=drawColor,anchor=base,inner sep=0pt, outer sep=0pt, scale=  0.89] at (102.11, 19.17) {(title-only)};

\node[text=drawColor,anchor=base,inner sep=0pt, outer sep=0pt, scale=  0.89] at (136.70, 28.78) {WS-TB};

\node[text=drawColor,anchor=base,inner sep=0pt, outer sep=0pt, scale=  0.89] at (171.30, 28.78) {DQG};
\end{scope}
\begin{scope}
\path[clip] (  0.00,  0.00) rectangle (187.90,108.41);
\definecolor{drawColor}{RGB}{0,0,0}

\path[draw=drawColor,line width= 0.4pt,line join=round,line cap=round] ( 48.56,  0.00) rectangle (155.66, 11.75);
\end{scope}
\begin{scope}
\path[clip] (  0.00,  0.00) rectangle (187.90,108.41);

\path[] ( 48.56, -0.00) rectangle (155.66, 11.75);
\end{scope}
\begin{scope}
\path[clip] (  0.00,  0.00) rectangle (187.90,108.41);
\definecolor{fillColor}{RGB}{255,255,255}

\path[fill=fillColor] ( 56.06,  1.00) rectangle ( 65.81, 10.75);
\end{scope}
\begin{scope}
\path[clip] (  0.00,  0.00) rectangle (187.90,108.41);
\definecolor{fillColor}{RGB}{0,114,178}

\path[fill=fillColor] ( 56.77,  1.71) rectangle ( 65.09, 10.03);
\end{scope}
\begin{scope}
\path[clip] (  0.00,  0.00) rectangle (187.90,108.41);
\definecolor{fillColor}{RGB}{255,255,255}

\path[fill=fillColor] (108.94,  1.00) rectangle (118.69, 10.75);
\end{scope}
\begin{scope}
\path[clip] (  0.00,  0.00) rectangle (187.90,108.41);
\definecolor{fillColor}{RGB}{86,180,233}

\path[fill=fillColor] (109.65,  1.71) rectangle (117.98, 10.03);
\end{scope}
\begin{scope}
\path[clip] (  0.00,  0.00) rectangle (187.90,108.41);
\definecolor{drawColor}{RGB}{0,0,0}

\node[text=drawColor,anchor=base west,inner sep=0pt, outer sep=0pt, scale=  0.88] at ( 71.31,  2.84) {TF$*$IDF};
\end{scope}
\begin{scope}
\path[clip] (  0.00,  0.00) rectangle (187.90,108.41);
\definecolor{drawColor}{RGB}{0,0,0}

\node[text=drawColor,anchor=base west,inner sep=0pt, outer sep=0pt, scale=  0.88] at (124.19,  2.84) {Jaccard};
\end{scope}
\end{tikzpicture}
     \caption{Average overlap of texts from positive training instances (words were stemmed and lowercased).}
    \label{fig:analysis:overlap}
\end{figure}
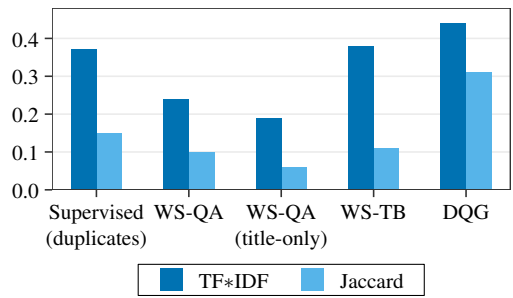

Figure \ref{fig:analysis:overlap} shows the Jaccard coefficient and the TF$*$IDF score averaged over all instances in the four StackExchange datasets of %
\S\ref{sec:results}. %
We observe that the overlap in WS-TB is similar to the overlap of actual duplicate questions in supervised training. The DQG overlap is higher, because generated titles only contain relevant content (e.g., no conversational phrases).
We also found that the BLEU scores of the MQAN model for QG are not very high 
(between 13.3--18.9 BLEU depending on the dataset), which shows that the texts are still sufficiently different.
The overlap shows that both our methods use suitable training data with sufficiently similar, but not fully redundant texts.

Interestingly, the overlap scores of question-answer pairs are lower, especially when considering title-answer pairs as it is the case in the answer selection experiments (\S\ref{sec:additional:answer-selection}). 
This could explain one factor that may contribute to the better scores that we achieve with WS-TB for BiLSTM in this scenario. Because the overlap of title-body pairs is higher, the siamese network can receive a stronger training signal for positive instances, which could lead to better representations for similarity scoring.

\subsection{Qualitative Analysis}
\begin{figure}

\centering

\begin{tikzpicture}[xscale=.77,yscale=.77]

\definecolor{lightgreen}{RGB}{205,245,195}
\definecolor{lightblue}{RGB}{228,232,252}
\definecolor{lightyellow}{RGB}{255,255,220}
\definecolor{lightgray}{RGB}{245,245,245}

\node(title1)[draw,align=left,anchor=north west,text width=7.4cm,font=\footnotesize,fill=lightgreen] 
{\textbf{Title:} 14.10, 15.04 - HDMI audio not working on Dell Vostro 3750 - nVidia card not detected by aplay -l};
\node(mqan1)[draw,below=-0.2mm of title1.south west,align=left,anchor=north west,text width=7.4cm,font=\footnotesize,fill=lightgray] 
{\textbf{DQG:} ALSA not detected in nVidia};

\node(title2)[draw,below=1.5mm of mqan1.south west,align=left,anchor=north west,text width=7.4cm,font=\footnotesize,fill=lightgreen] 
{\textbf{Title:} Installing ubuntu 12.04.02 in uefi mode};
\node(mqan2)[draw,below=-0.2mm of title2.south west,align=left,anchor=north west,text width=7.4cm,font=\footnotesize,fill=lightgray] 
{\textbf{DQG:} Ubuntu 16.04 LTS boot loader not working};

\node(title3)[draw,below=1.5mm of mqan2.south west,align=left,anchor=north west,text width=7.4cm,font=\footnotesize,fill=lightgreen] 
{\textbf{Title:} Grub2 not updating};
\node(mqan3)[draw,below=-0.2mm of title3.south west,align=left,anchor=north west,text width=7.4cm,font=\footnotesize,fill=lightgray] 
{\textbf{DQG:} How to fix Grub2 error};

\end{tikzpicture}
\caption{Random samples of titles and DQG output.}
\label{fig:analysis:example-1}
\end{figure}

To better understand DQG, we manually inspected a random sample of 200 generated questions and title-body pairs from multiple platforms. 
Three titles and generated duplicates from AskUbuntu are shown in Figure \ref{fig:analysis:example-1}.

We found that most of the generated duplicates are sensible, 
and most of the error cases fall into one of the following two categories:

(1) Some generated questions are off-topic because they contain information that was generated from a body with minimal overlap to the title (see example 4 in the Appendix). 

(2) Some questions include wrong version numbers or wrong names (see example 5 in the Appendix, or the second example in Figure \ref{fig:analysis:example-1}). %

However, we also found that many of the generated titles contain complementary information to the original title, which can be seen in Figure \ref{fig:analysis:example-1} (e.g., `ALSA', `boot loader' etc).

%

%
We also inspected the training data of the difficult DQG domain transfer case to explore reasons for the small performance decreases when transferring the question generation model from StackExchange Academia to technical domains. Most importantly, we find that the model often falls back to copying important phrases from the body and sometimes generates additional words from the source domain. We note that this is not the case for models without copy mechanisms, e.g., Transformer often generates unrelated text %
(examples are in the Appendix).

\section{Conclusion}

In this work, we have %
trained
duplicate question detection models without labeled training data. This can be beneficial for a large number of \cqa forums that do not contain enough annotated duplicate questions or question-answer pairs to use existing training methods. %
Our two novel methods, duplicate question generation and weak supervision with title-body pairs, use only information of unlabeled questions and can thus utilize 
more training data. 
While both are already very effective 
with 
 the same number of training instances as compared to other methods (e.g., outperforming adversarial domain transfer), our experiments have shown that we can outperform even supervised training when using larger amounts of 
 unlabeled questions. %

Further, we have demonstrated that weak supervision with title-body pairs is well-suited to train answer selection models without direct answer supervision.
This shows that our work can potentially benefit a much wider range of related tasks beyond duplicate question detection.
For instance, future work could use our methods to obtain training data from \cqa forums in several languages, potentially improving cross-lingual \cqa \cite{Joty2017,rueckle:WWW:2019}. It could also combine the different training methods to achieve further improvements. 

Our source code and data is publicly available.\footnote{\url{http://github.com/UKPLab/emnlp2019-duplicate_question_detection}}

\section*{Acknowledgements}
This work has been supported by the German Research Foundation (DFG) as part of the
QA-EduInf project (grant GU 798/18-1 and grant RI 803/12-1),
by the German Federal Ministry of Education and Research (BMBF) under the promotional reference 03VP02540 (ArgumenText),
and
by the German Federal Ministry of Education and Research (BMBF) 
as part of the Software Campus program under the promotional reference 01IS17050.

\bibliography{main}
\bibliographystyle{acl_natbib}

\appendix

\newpage
\section{Supplemental Material}
\label{sec:supplemental}

\subsection{Filtering and Paragraph Extraction}
Filtering and paragraph selection is necessary to obtain less noisy title-body pairs for QG. This consists of three steps:
(1) filtering out questions that are not suitable for QG, 
(2) extracting paragraphs from bodies, 
and (3) only keeping one paragraph with the highest similarity to the title. 
The details are given below.

\paragraph{Filtering.}
We discard all questions that:
\begin{itemize}[noitemsep]
    \item contain bodies with less than 10 words,
    \item are downvoted, i.e., have a score on StackExchange that is below zero (`bad' questions).
\end{itemize}

\paragraph{Paragraph extraction.}
Some questions contain multiple long paragraphs, which is too much information to train suitable question generation or duplicate detection models. We thus extract paragraphs from the text to filter them in a later step.

In StackExchange platforms, users can freely add new lines, new paragraphs (the text then appears in HTML paragraph tags), lists, images, and code. This freedom results in many different ways of writing text. For instance, some users prefer to use paragraph tags and other users separate every sentence with a new-line character and all paragraphs with two or more new-line characters. Further, many users include code and enumerations in their questions. 

This makes it difficult to extract actual paragraphs of the text. Thus, we first apply a preprocessing step to remove all HTML tags:
\begin{itemize}[noitemsep]
    \item We remove all code and images from the description.
    \item We then extract the text of each item from enumerations and append a new-line character. 
    \item Likewise, we extract the text in paragraph tags and append a new-line character. We retain all new-line characters that appear in the paragraph.
\end{itemize}

We then analyze the new-line characters in the text to form the paragraphs for extraction. We read the input line-by-line:
\begin{itemize}[noitemsep]
    \item If the current line contains only one sentence it is merged with the previous paragraph.
    \item If the current line contains more than one sentence it is considered as a new paragraph.
\end{itemize}

\paragraph{Paragraph selection.}
After extracting N paragraphs $p_1 \ldots p_N$ from the description, we 
select one paragraph according to $\argmax_{p_n} f(p_n, \mathrm{title}(q))$. 
The function $f$
scores each $p_n$ by calculating the maximum cosine similarity of a sentence s in $p_n$ to the question title $\mathrm{title}(q)$ using a sentence encoder ($\mathit{enc}$):
\begin{equation*}
    f(p_i, t) = \max_{s \in p_i} \left[ \cos(\mathrm{enc}(s), \mathrm{enc}(t)) \right]
\end{equation*}
In our experiments, $\mathrm{enc}$ is 
the (monolingual) %
encoder
of \citet{rueckle:2018}, which uses different pooling strategies with multiple types of word embeddings.
We calculate the maximum similarity of individual sentences to determine the semantic similarity independent of the paragraph length.

\subsection{DQG with the Transformer}

\begin{table*}
    \centering
    \footnotesize
    \begin{tabular}{l|r|rrrrr}
        \toprule
         & \textbf{AskUbuntu-Lei} & \textbf{Android} & \textbf{Apple} & \textbf{AskUbuntu} & \textbf{SuperUser} & Average \\
        \multicolumn{2}{r}{Measuring P@5. Results (dev / test) for RCNN} & \multicolumn{5}{|c}{Measuring AUC(0.05). Results for BiLSTM} \\
        \midrule
        \multicolumn{6}{l}{\textbf{Trained on 1x data}~~\scriptsize{(all methods use the same number of training instances as in supervised training)}} \\
        \midrule
        Supervised (in-domain) & 48.0 / 45.0 & - & - & 0.848 & 0.944 & - \\
        Adversarial Transfer (best) & - & 0.790 & 0.861 & 0.796 & 0.911 & 0.840 \\
        DQG w. MQAN & 46.4 / 44.8 & 0.793 & 0.870 & 0.801 & 0.921 & 0.846 \\
        DQG w. Transformer & 47.2 / 44.9 & 0.723 & 0.809 & 0.799 & 0.917 & 0.812 \\
        WS-TB & 46.4 / 45.4 & 0.811 & 0.866 & 0.804 & 0.913 & 0.849 \\
        \midrule
        \multicolumn{6}{l}{\textbf{Trained on all available data}} \\
        \midrule
        DQG w. MQAN & 47.4 / 44.3 & 0.833 & 0.911 & 0.855 & 0.944 & 0.886 \\
        DQG w. Transformer & 46.4 / 44.7 & 0.783 & 0.876 & 0.836 & 0.942 & 0.859 \\        
        WS-TB & 47.3 / 45.3 & 0.852 & 0.910 & 0.871 & 0.952 & 0.896 \\
        \bottomrule
    \end{tabular}
    \caption{Results of the models with different training strategies, including DQG with the Transformer.}
    \label{tbl:appendix:results}
\end{table*}

\begin{table*}
    \centering
    \footnotesize
    \begin{tabular}{ll|rr|rrrr}
        \textbf{Source} & \textbf{Target} & \multicolumn{2}{|c}{\textbf{Domain Transfer}} & \multicolumn{4}{|c}{\textbf{Duplicate Question Generation}} \\
         & & Direct & Adversarial & Transformer & $\Delta$ & MQAN & $\Delta$\\
        \toprule
        \multirow{3}{*}{AskUbuntu} 
         & Android & 0.692 & 0.790 & 0.762 & $+$0.039 & \textbf{0.797} & $+$0.004\\
         & Apple & 0.828 & 0.855 & 0.821 & $+$0.012 & \textbf{0.861} & $-$0.009\\
         & SuperUser & 0.908 & 0.911 & 0.913 & $-$0.004 & \textbf{0.916} & $-$0.005 \\
         \midrule
        \multirow{3}{*}{SuperUser} 
         & Android & 0.770 & 0.790 & 0.755 & $+$0.032 & \textbf{0.794} & $+$0.001 \\
         & Apple & 0.828 & \textbf{0.861} & 0.833 & $+$0.024 & \textbf{0.861} & $-$0.009 \\
         & AskUbuntu & 0.730 & 0.796 & 0.797 & $-$0.002 & \textbf{0.809} & $+$0.008 \\
         \bottomrule
    \end{tabular}
    \caption{Domain transfer performances including Transformer. $\Delta$ denotes the difference to the setup with in-domain DQG.}
    \label{tbl:appendix:transfer}
\end{table*}

\begin{table}
    \centering
    \footnotesize
    \begin{tabular}{ll|rr}
        \textbf{Source} & \textbf{Target} & \multicolumn{2}{|c}{\textbf{DQG}} \\
         & & Transformer & MQAN \\
        \toprule
        \multirow{3}{*}{Travel} 
         & Android & 0.550 & 0.789 \\
         & Apple & 0.624 & 0.864 \\ 
         & SuperUser & 0.856 & 0.914 \\
         & AskUbuntu & 0.664 & 0.787 \\
                 \midrule
        \multirow{3}{*}{Academia} 
         & Android & 0.530 & 0.776 \\
         & Apple & 0.576 & 0.854 \\ 
         & SuperUser & 0.840 & 0.912 \\
         & AskUbuntu & 0.672 & 0.760 \\
         \bottomrule
    \end{tabular}
    \caption{The DQG domain transfer performance of different question generation models from more distant source domains that offer smaller numbers of unlabelled questions.}
    \label{tbl:appendix:transfer-distant}
\end{table}

In addition to MQAN \cite{McCann2018decaNLP}, we also experimented with the Transformer \cite{Vaswani2017} for question generation using the Tensor2Tensor library \cite{tensor2tensor}. The most notable difference to MQAN is that the Transformer does not include a copy mechansim. 

For our experiments we use the official implementation of the Transformer and use the same encoder-decoder approach as in machine translation. But instead of translating an input sentence to a target language, we generate a question from a paragraph of the body.

The results are given in Table \ref{tbl:appendix:results}. We observe that Transformer performs worse than MQAN in domains that offer fewer unlabeled questions (Android, Apple). In contrast, for domains with more unlabeled questions (AskUbuntu, SuperUser), DQG with Transformer performs on the same level or only mildly worse than DQG with MQAN.

We also tested Transformer in the domain transfer scenarios. Table \ref{tbl:appendix:transfer} shows the results when transferring from close domains, and Table \ref{tbl:appendix:transfer-distant} shows the results when transferring from more distant domains.
In contrast to MQAN, the performance of Transformer substantially decreases. 
We observe that MQAN is much more robust against domain changes due to its copy mechanism, which allows it to copy words and phrases from the input text. In contrast, Transformer falls back to outputting unrelated (but grammatical) domain-specific text. 
 Examples are given in Appendix \ref{ap:examples} below.

Thus, different QG models can have a substantial impact on the performance of DQG. However, this also suggests that better models could have a positive effect on DQG performance, potentially improving upon DQG with MQAN.
\subsection{BERT Setup}
\label{ap:bert}

For our experiments in Section \ref{sec:additional:bert} we add BERT to two experimental frameworks. In both extensions we use the HuggingFace implementation\footnote{\url{https://github.com/huggingface/pytorch-transformers}}.

We add BERT as a sentence encoder to the experimental software of \cite{Shah2018} and average over all BERT output states to obtain question representations. The rest of the implementation is the same as for BiLSTM (e.g., loss calculation). We train the models until they do not improve for at least 20 epochs, and we restore the weights of the epoch that obtained the best development score.

For all other datasets (AskUbuntu-Lei and Answer Selection datasets) we add BERT to the experimental software of \newcite{rueckle:AAAI:2019}. We do not include it in the software of \newcite{Lei2016} because it is tightly coupled to the Theano framework, which is not actively maintained. We add BERT as a pairwise classification model which then directly scores question-question pairs and question-answer pairs (binary labels for training). The output prediction is used as a score for ranking. We train the models for 10 epochs and restore the weights of the epoch with the best development score. %

\subsection{BERT Results}
\label{ap:bert-details}

In \S\ref{sec:additional:answer-selection} we report averaged scores over the five StackExchange answer selection datasets of \cite{rueckle:AAAI:2019}. We include the individual scores in Table~\ref{tbl:ap:results:bert:as}.

Additionally, we provide the MAP, MRR, and P@1 (accuracy) scores for AskUbuntu-Lei in Table~\ref{tbl:ap:results:bert:au}. These are averaged over five runs due to the small dev/test sizes.

\begin{table*}
    \centering
    \footnotesize
    \begin{tabular}{l|rrrrr}
        \toprule & \textbf{Travel} & \textbf{Apple} & \textbf{Aviation} & \textbf{Academia} & \textbf{Cooking} \\
        \midrule
        Supervised (in-domain) & 67.8 & 43.8 & 62.6 & 53.9 & 55.9 \\
        WS-TB (1x) & 62.3 & 42.8 & 65.5 & 54.2 & 52.8\\
        WS-TB (8x) & 69.5 & 47.3 & 64.9 & 58.7 & 58.3 \\
        \bottomrule
    \end{tabular}
    \caption{Accuracy scores of BERT on the individual answer selection datasets.}
    \label{tbl:ap:results:bert:as}
\end{table*} 

\begin{table*}
    \centering
    \footnotesize
    \begin{tabular}{l|rrrr}
        \toprule & \textbf{P@1} & \textbf{P@5} & \textbf{MAP} & \textbf{MRR} \\
        \midrule
        Supervised (in-domain) & 69.25 & 52.30 & 69.13 & 80.35 \\
        WS-TB (1x) & 61.08 & 47.25 & 61.81 & 74.53 \\
        WS-TB (8x) & 65.70 & 49.61 & 64.58 & 77.98 \\
        \bottomrule
    \end{tabular}
    \caption{Results of BERT on AskUbuntu-Lei.}
    \label{tbl:ap:results:bert:au}
\end{table*}


\subsection{Additional QG Examples}
\label{ap:examples}
Below we show examples of generated questions. 
The questions generated with MQAN more closely retain the meaning of the body or paragraph, but Transformer questions also contain the relevant keywords (except for the transfer cases).
Examples 4 and 5 refer to the error cases mentioned in our analysis (see \S\ref{sec:analysis}).
\\[5mm]
\begin{tikzpicture}[xscale=.77,yscale=.77]

\definecolor{lightgreen}{RGB}{205,245,195}
\definecolor{lightblue}{RGB}{228,232,252}
\definecolor{lightyellow}{RGB}{255,255,220}
\definecolor{lightgray}{RGB}{245,245,245}

\node(headline)[draw=none,align=left,anchor=north west] at (0,8) 
{\textbf{Example 1}};

\node(titleheader)[draw=none,below=8mm of headline.south,align=left,anchor=north west,font=\scriptsize] at (0,8) 
{\textbf{QUESTION}};

\node(title)[draw,below=0.5mm of titleheader.south west,align=left,anchor=north west,text width=7.4cm,font=\footnotesize,fill=lightgreen] 
{how to get beep working?};

\node(bodyheader)[draw=none,below=1.5mm of title.south west,align=left,anchor=north west,font=\scriptsize]
{\textbf{RELEVANT PARAGRAPH}};

\node(body)[draw,below=0.5mm of bodyheader.south west,align=left,anchor=north west,text width=7.4cm,font=\footnotesize,fill=lightblue] 
{I have a laptop, i installed the "beep" package. I turned every sound to full, and i: but i can't hear any "beeping" sound. What am I missing? I just need to run the "beep" when a script is finished. Thank you for any links/howtos!};

\draw[-] (title.south) to[out=270,in=90] (body.north);

\node(mqandup_head)[draw=none,below=2mm of body.south west,align=left,anchor=north west,font=\scriptsize]
{\textbf{IN-DOMAIN QG MODELS}};
\node(mqandup)[draw,below=0.5mm of mqandup_head.south west,align=left,anchor=north west,text width=7.4cm,font=\footnotesize,fill=lightblue] 
{\textbf{MQAN:} How to fix beep package?};
\node(transformerdup)[draw,below=1mm of mqandup.south west,align=left,anchor=north west,text width=7.4cm,font=\footnotesize,fill=lightblue] 
{\textbf{Transformer:} How to remove `` beep '' from my laptop?};

\node(placeholder)[draw=none,below=7cm of transformerdup.south,align=left,anchor=south west,font=\scriptsize]{};

\end{tikzpicture}
\\[5mm]
\begin{tikzpicture}[xscale=.77,yscale=.77]

\definecolor{lightgreen}{RGB}{205,245,195}
\definecolor{lightblue}{RGB}{228,232,252}
\definecolor{lightyellow}{RGB}{255,255,220}
\definecolor{lightgray}{RGB}{245,245,245}

\node(headline)[draw=none,align=left,anchor=north west] at (0,8) 
{\textbf{Example 2}};

\node(titleheader)[draw=none,below=8mm of headline.south,align=left,anchor=north west,font=\scriptsize] at (0,8) 
{\textbf{QUESTION}};

\node(title)[draw,below=0.5mm of titleheader.south west,align=left,anchor=north west,text width=7.4cm,font=\footnotesize,fill=lightgreen] 
{13" MacBook Pro with Win 7 and External VGA gets 640x480};

\node(bodyheader)[draw=none,below=1.5mm of title.south west,align=left,anchor=north west,font=\scriptsize]
{\textbf{RELEVANT PARAGRAPH}};

\node(body)[draw,below=0.5mm of bodyheader.south west,align=left,anchor=north west,text width=7.4cm,font=\footnotesize,fill=lightblue] 
{I have a brand new 13'' MacBook Pro - 2.26 GHz and the NVIDIA 9400M Video card. I installed Windows 7 (final) in boot camp and booted up to Windows 7. Installed all the drivers from the Apple disk and it was working great. Then I attached the external VGA adapter (from apple) to connect to a projector and it dropped down at 640x480 resolution. No matter what I did it would n't let me change to a higher resolution if the external VGA was connected. Once it disconnects then it goes back to the normal resolution. If I am booted into Snow Leopard it works fine. I tried updating the NVIDIA drivers and it behaved exactly the same. Ultimately I want to get 1024x768 or better resolution when connected to an external display. If it isn't fixable then I am curious if anyone else has seen this, if it is a known issue, and who to contact for support (Apple, Microsoft or NVIDIA?) Update: Just attaching the Mini-DVI to VGA adapter kicks it into 640x480, no projector is required. I tried forcing the display driver from Generic PnP Monitor to one that supported 1024x768 and that didn't work either.};

\draw[-] (title.south) to[out=270,in=90] (body.north);

\node(mqandup_head)[draw=none,below=2mm of body.south west,align=left,anchor=north west,font=\scriptsize]
{\textbf{IN-DOMAIN QG MODELS}};
\node(mqandup)[draw,below=0.5mm of mqandup_head.south west,align=left,anchor=north west,text width=7.4cm,font=\footnotesize,fill=lightblue] 
{\textbf{MQAN:} 1024x768 or better resolution when connected to an external display};
\node(transformerdup)[draw,below=1mm of mqandup.south west,align=left,anchor=north west,text width=7.4cm,font=\footnotesize,fill=lightblue] 
{\textbf{Transformer:} How can I get an external display to work?};

\end{tikzpicture}
\begin{tikzpicture}[xscale=.77,yscale=.77]

\definecolor{lightgreen}{RGB}{205,245,195}
\definecolor{lightblue}{RGB}{228,232,252}
\definecolor{lightyellow}{RGB}{255,255,220}
\definecolor{lightgray}{RGB}{245,245,245}

\node(headline)[draw=none,align=left,anchor=south west,text width=7.4cm] at (0,8) 
{\textbf{Example 3}.
This contains a particularly short description with not much overlap to the question. Such examples could be removed with additional preprocessing / filtering.};

\node(titleheader)[draw=none,below=3mm of headline.south,align=left,anchor=north west,font=\scriptsize] at (0,8) 
{\textbf{QUESTION}};

\node(title)[draw,below=0.5mm of titleheader.south west,align=left,anchor=north west,text width=7.4cm,font=\footnotesize,fill=lightgreen] 
{How to echo contents of file in a DOS/Windows command prompt?};

\node(bodyheader)[draw=none,below=1.5mm of title.south west,align=left,anchor=north west,font=\scriptsize]
{\textbf{RELEVANT PARAGRAPH}};

\node(body)[draw,below=0.5mm of bodyheader.south west,align=left,anchor=north west,text width=7.4cm,font=\footnotesize,fill=lightblue] 
{Like in Unix' cat filename ... Can this be done?};

\draw[-] (title.south) to[out=270,in=90] (body.north);

\node(mqandup_head)[draw=none,below=2mm of body.south west,align=left,anchor=north west,font=\scriptsize]
{\textbf{IN-DOMAIN QG MODELS}};
\node(mqandup)[draw,below=0.5mm of mqandup_head.south west,align=left,anchor=north west,text width=7.4cm,font=\footnotesize,fill=lightblue] 
{\textbf{MQAN:} How to make cat filename in Unix' cat filename ?};
\node(transformerdup)[draw,below=1mm of mqandup.south west,align=left,anchor=north west,text width=7.4cm,font=\footnotesize,fill=lightblue] 
{\textbf{Transformer:} What is the difference between 'cat' and `cat` in Unix?};

\end{tikzpicture}
\\[15mm]
\begin{tikzpicture}[xscale=.77,yscale=.77]

\definecolor{lightgreen}{RGB}{205,245,195}
\definecolor{lightblue}{RGB}{228,232,252}
\definecolor{lightyellow}{RGB}{255,255,220}
\definecolor{lightgray}{RGB}{245,245,245}

\node(headline)[draw=none,align=left,anchor=south west,text width=7.4cm] at (0,8) 
{\textbf{Example 4} (from Analysis).
The duplicates were generated based on a paragraph that does not have much overlap with the question.};

\node(titleheader)[draw=none,below=3mm of headline.south,align=left,anchor=north west,font=\scriptsize] at (0,8) 
{\textbf{QUESTION}};

\node(title)[draw,below=0.5mm of titleheader.south west,align=left,anchor=north west,text width=7.4cm,font=\footnotesize,fill=lightgreen] 
{How can I remove an autostart service that is not listed in gnome-session-properties?};

\node(bodyheader)[draw=none,below=1.5mm of title.south west,align=left,anchor=north west,font=\scriptsize]
{\textbf{RELEVANT PARAGRAPH}};

\node(body)[draw,below=0.5mm of bodyheader.south west,align=left,anchor=north west,text width=7.4cm,font=\footnotesize,fill=lightblue] 
{Today I upgraded from raring to saucy. This brought the Ubuntu One icon back to my indicator applet, which I had disabled. So I removed the line NoDiplay=true from /etc/xdg/autostart/ubuntuone-launch.desktop. But still Ubuntu One doesn't show up in gnome-session-properties and I can't disable it. What gives?};

\draw[-] (title.south) to[out=270,in=90] (body.north);

\node(mqandup_head)[draw=none,below=2mm of body.south west,align=left,anchor=north west,font=\scriptsize]
{\textbf{IN-DOMAIN QG MODELS}};
\node(mqandup)[draw,below=0.5mm of mqandup_head.south west,align=left,anchor=north west,text width=7.4cm,font=\footnotesize,fill=lightblue] 
{\textbf{MQAN:} Ubuntu One doesn't show up in gnome-session-properties};
\node(transformerdup)[draw,below=1mm of mqandup.south west,align=left,anchor=north west,text width=7.4cm,font=\footnotesize,fill=lightblue] 
{\textbf{Transformer:} Why doesn't Ubuntu One sync with an indicator?};

\end{tikzpicture}

\vspace{10mm}

\begin{tikzpicture}[xscale=.77,yscale=.77]

\definecolor{lightgreen}{RGB}{205,245,195}
\definecolor{lightblue}{RGB}{228,232,252}
\definecolor{lightyellow}{RGB}{255,255,220}
\definecolor{lightgray}{RGB}{245,245,245}

\node(headline)[draw=none,align=left,anchor=south west,text width=7.4cm] at (0,8) 
{\textbf{Example 5} (from Analysis).
The question generated by the (in-domain) Transformer model is suitable, but it does not contain the correct product name of the printer (``0b'' instead of ``LBP2900b''). However, even the MQAN model that was trained on StackExchange Travel is able to correctly copy all necessary information from the input. The Transformer trained on StackExchange Travel fails with generic (and grammatical) text from the travel domain.};

\node(titleheader)[draw=none,below=3mm of headline.south,align=left,anchor=north west,font=\scriptsize] at (0,8) 
{\textbf{QUESTION}};

\node(title)[draw,below=0.5mm of titleheader.south west,align=left,anchor=north west,text width=7.4cm,font=\footnotesize,fill=lightgreen] 
{How to install Canon LBP2900b drivers?};

\node(bodyheader)[draw=none,below=1.5mm of title.south west,align=left,anchor=north west,font=\scriptsize]
{\textbf{RELEVANT PARAGRAPH}};

\node(body)[draw,below=0.5mm of bodyheader.south west,align=left,anchor=north west,text width=7.4cm,font=\footnotesize,fill=lightblue] 
{I am trying very hard to install Canon LBP2900b Printer in Ubuntu 13.10. I have searched and googled a lot for the solution over fortnight but none of the site / link gave me the simple solution for me. How can accomplish my goal?};

\draw[-] (title.south) to[out=270,in=90] (body.north);

\node(mqandup_head)[draw=none,below=2mm of body.south west,align=left,anchor=north west,font=\scriptsize]
{\textbf{IN-DOMAIN QG MODELS}};
\node(mqandup)[draw,below=0.5mm of mqandup_head.south west,align=left,anchor=north west,text width=7.4cm,font=\footnotesize,fill=lightblue] 
{\textbf{MQAN:} How to install Canon LBP2900b Printer in Ubuntu 13.10?};
\node(transformerdup)[draw,below=1mm of mqandup.south west,align=left,anchor=north west,text width=7.4cm,font=\footnotesize,fill=lightblue] 
{\textbf{Transformer:} How to Install Canon 0b Printer on Ubuntu 13.10?};

\node(mqandup_t_head)[draw=none,below=2mm of transformerdup.south west,align=left,anchor=north west,font=\scriptsize]
{\textbf{DOMAIN TRANSFER QG MODELS} (from SE Travel)};
\node(mqandup_t)[draw,below=0.5mm of mqandup_t_head.south west,align=left,anchor=north west,text width=7.4cm,font=\footnotesize,fill=lightblue] 
{\textbf{MQAN:} How to install to install Canon LBP2900b in Ubuntu 13.10?};
\node(transformerdup_t)[draw,below=1mm of mqandup_t.south west,align=left,anchor=north west,text width=7.4cm,font=\footnotesize,fill=lightblue] 
{\textbf{Transformer:} How can I find my boat in Hokkaido?};

\end{tikzpicture}

\end{document}